\documentclass[10pt,twocolumn,letterpaper]{article}

\usepackage[pagenumbers]{cvpr} 

\definecolor{mygray}{gray}{0.94}
\definecolor{lightgray}{gray}{0.75}
\definecolor{first}{rgb}{1,0.7,0.7}
\definecolor{second}{rgb}{1,0.85,0.7}
\definecolor{third}{rgb}{1,1,0.8}

\usepackage[most]{tcolorbox}
\usepackage{setspace}

\newcommand{\cmark}{\ding{51}}  
\newcommand{\xmark}{\ding{55}}  
\definecolor{cvprblue}{rgb}{0.21,0.49,0.74}
\definecolor{lightgreen}{RGB}{220, 255, 220}
\usepackage[pagebackref,breaklinks,colorlinks,allcolors=cvprblue]{hyperref}
\usepackage{booktabs}
\usepackage{multirow}
\usepackage{graphicx} 
\usepackage[table]{xcolor}
\usepackage{arydshln} %
\usepackage{pifont}
\def\confName{CVPR}
\def\confYear{2026}

\title{EgoProx: Evaluating MLLMs on Egocentric 3D Proximity Reasoning \\ Across a Cognitive Hierarchy}

\author{
Jinzhao Li$^{1,2}$, 
Yinuo Chen$^{1*}$, 
Dongxu Piao$^{1*}$, 
Panwang Pan$^{2\dag}$, 
Yifan Yu$^{2}$, 
Dong Wang$^{2}$, 
Honglei Yan$^{2}$, \\
Liang Yue$^{1}$, 
Shaofei Wang$^{3}$, 
Yixin Chen$^{3}$, 
Siyuan Huang$^{3}$, 
Miao Liu$^{1\ddagger}$ \\
$^{1}$College of AI, Tsinghua University \\
$^{2}$ByteDance \\
$^{3}$State Key Laboratory of General Artificial Intelligence, BIGAI \\
{\tt\small \url{https://lijinzhao30.github.io/Egoprox/}}
}

\begin{document}

\twocolumn[{
\renewcommand\twocolumn[1][]{#1}
\maketitle
\begin{center}
\vspace{-0.5em}
\includegraphics[width=\linewidth]{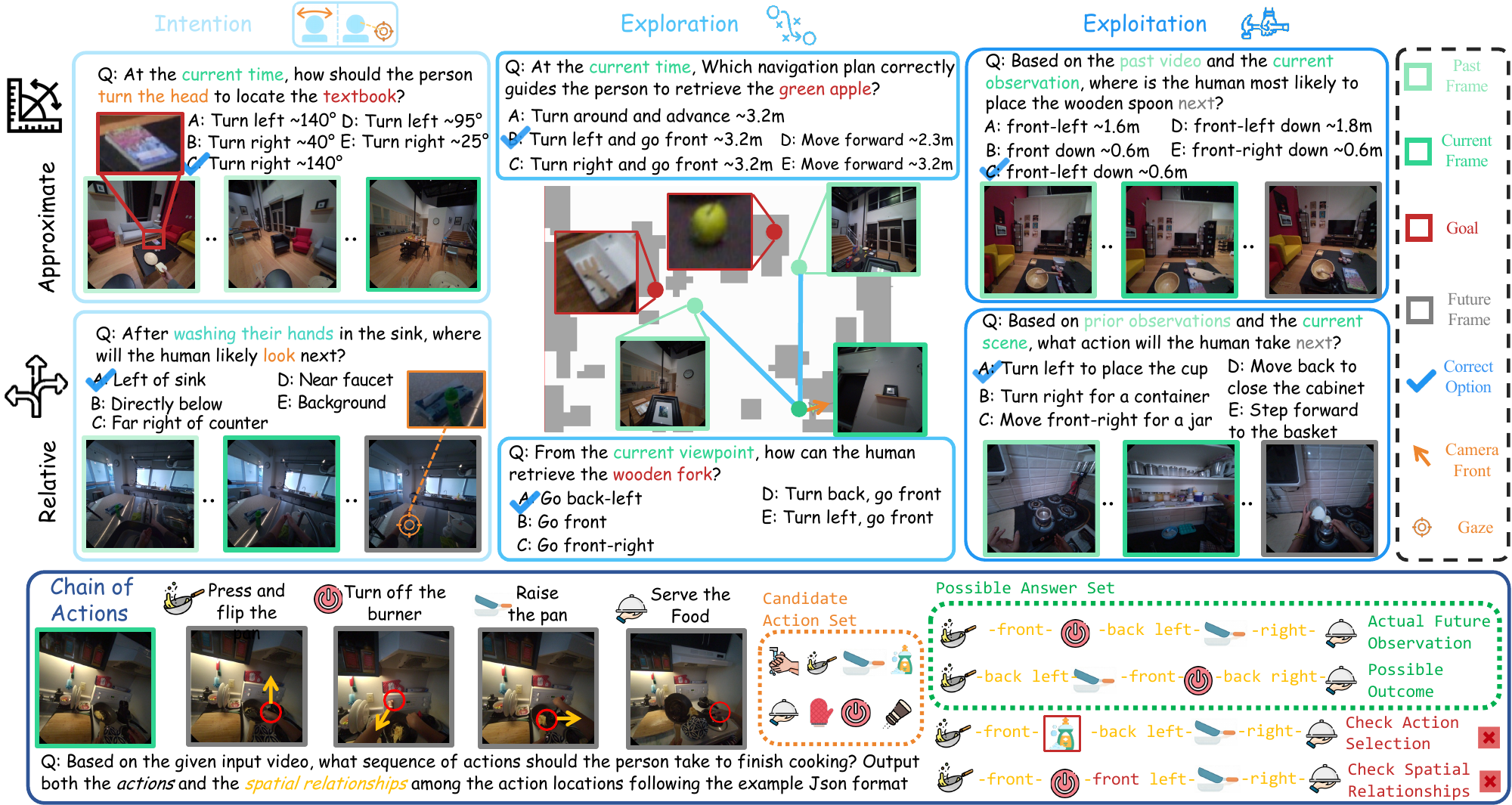}
\vspace{-1.5em}
\captionof{figure}{\textbf{Visual illustration of the EgoProx benchmark}. We aim to evaluate multimodal large language models (MLLMs) on complex egocentric proximity reasoning tasks that require 4D action and scene understanding. Our benchmark spans four core dimensions following a cognitive hierarchy: Intention, Exploration, Exploitation, and Chain of Actions. We adopt approximate transformations and relative spatial relationships to represent proximity. The examples illustrate the model’s need to interpret long-term contextual cues, spatial dependencies, and action-state changes from first-person visual inputs, providing a comprehensive assessment of egocentric spatial intelligence.}
\label{fig:teaser}
\end{center}
}]

\begingroup
\renewcommand\thefootnote{}
\footnotetext{*Equal contribution. $\dag$ Project Lead. $\ddagger$ The corresponding author.}
\endgroup

\begin{abstract}
Humans constantly reason about 3D proximity, the relations between their body and surrounding objects, to guide perception and action in daily life. Whether multimodal large language models (MLLMs) can perform such embodied 3D reasoning remains unclear. To this end, we introduce \textbf{EgoProx}, a benchmark for egocentric 3D proximity reasoning. We organize our tasks along a cognitive chain, covering intention, exploration, exploitation, and chain-of-actions reasoning. We also design an agent based data engine that produces diverse and consistent QA pairs at scale. We benchmark prevailing MLLMs on EgoProx and conduct additional analyses with dataset specific and task specific instruction tuning. We observe large cross-domain gains, indicating that current MLLMs contain some spatial knowledge; however, they still struggle to effectively leverage it for spatial reasoning VQA.


\end{abstract}    
\section{Introduction}
\label{sec:intro}

Humans constantly reason about 3D proximity, the spatial relations between their body and nearby objects in everyday life. Through 3D spatial awareness, the cognitive system drives intention such as head orientation and gaze shifts, leading to coordinated motor behaviors like locomotion and reaching, which further support hierarchical interactions in complex 3D scenes~\cite{bellmund2018navigating}. This 3D reasoning capability is a key mechanism that connects perception and actions. However, despite rapid advances in multimodal large language models (MLLMs) ~\cite{radford2021clip, jia2021align, li2022blip, alayrac2022flamingo, wang2022git, chen2022pali, li2023blip2, huang2023language, zhu2023minigpt, liu2023llava, dai2023instructblip, bai2023qwenvl, laurencon2023idefics, openai2023gpt4, ye2024mplug, chen2024internvl, wang2024qwen2vl, sun2024generative, deepmind2024gemini, wei2025deepseekocr}, it remains unclear whether current systems can emulate this human spatial reasoning within 3D scenes.

Egocentric video offers a natural lens for studying this problem. Its first-person viewpoint, embodiment, and continuous streaming reveal how humans form intentions through preparatory cues such as gaze and head motion, explore their surroundings, exploit spatial affordances, and ultimately coordinate chains of actions within 3D space~\cite{ma2016goingdeeper,akkil2016gaze,ma2024egocentric,peirone2024backpack,nunez2022egocentric}. An MLLM capable of reasoning about spatial proximity from the user’s perspective holds strong potential for applications in smart glasses, augmented reality, and robotics~\cite{Suzuki_2022, zha2025enablellm3dcapacity}.

However, despite the growing interest in egocentric MLLMs~\cite{lin2022egovlp,pramanick2023egovlpv2,zhao2023learning,suglia2024alanavlm,yang2025egolife,konrad2024gazegpt,xu2024retrieval,ohkawa2025exo2egodvc,hou2023groundnlq,wang2023lifelongmemory,kurita2023refego,sun2025intentiongrounding,lai2024lego,min2021film}, 3D proximity reasoning remains unexplored in existing egocentric visual question answering benchmarks. Establishing such a benchmark is essential to advance research in embodied spatial intelligence and to enable more capable AI systems. To this end, we introduce \textbf{Ego}centric \textbf{Prox}imity Reasoning (\textbf{EgoProx}), the first benchmark for assessing whether MLLMs can model the 3D perception–action coupling from a first-person perspective.

Here, we draw an analogy to the exploration and exploitation trade-off in machine learning. Unlike machine learning systems that must balance between exploration and exploitation, the egocentric viewpoint inherently captures how humans both explore and exploit the 3D world within a unified perceptual stream, while simultaneously encoding intention as the driver of embodied behavior. Consequently, we characterize 3D proximity reasoning along a cognitive hierarchy comprising three domains: \emph{intention}, \emph{exploration}, and \emph{exploitation}. As for the proximity measurements, we consider \emph{approximate proximity}, capturing metric transformations such as translation and rotation, and \emph{relative} proximity, describing spatial relationships between entities. Both reflect how humans naturally perceive spatial awareness. We further introduce a \emph{chain-of-actions} setting that extends our benchmark to assess higher-order cognitive processes underlying continuous human behavior in complex 3D scenes. We provide a visual illustration of our benchmark in Fig.~\ref{fig:teaser}.

A key challenge in constructing such a benchmark is designing a semi-automatic pipeline that supports VQA data generation. Unlike prior VQA benchmarks that rely on MLLMs with human-in-the-loop refinement~\cite{hou2023groundnlq, Mangalam2023EgoSchemaAD}, existing models lack the spatial intelligence to produce high-quality question–answer pairs~\cite{liu2023llava, zhu2023minigpt, bai2023qwenvl, ye2024mplug, zha2025enablellm3dcapacity}. Moreover, our diverse set of tasks require different reasoning capabilities, making a single foundation model insufficient. To address this, we develop an agent-based data engine that orchestrates multiple specialized tools to generate high-quality VQA data across diverse task types. Our agentic data engine tailors its workflow to the data generation requirements of each task type in our benchmark, it first applies the salient clip sampler to extract informative segments from long egocentric videos, and then selects and composes the appropriate tools from the 3D analysis toolset to complete VQA generation.Our key contributions are summarizes as follows:
\begin{itemize}
    \item We propose EgoProx, the first benchmark designed to evaluate whether MLLMs can reason 3D perception–action coupling from an egocentric point-of-view, with four tasks organized along a cognitive hierarchy: Intention, Exploration, Exploitation, and Chain of Actions. 
    \item We develop an agent-based data generation pipeline that leverages task-aware salient clip sampler and 3D analysis toolset to automatically synthesize high-quality VQA data across diverse task categories.
    
    \item Through extensive evaluation and cross-domain instruction-tuning experiments, we demonstrate that existing MLLMs already contain latent spatial knowledge acquired during pretraining, but unlocking this capability requires structured supervision. 
\end{itemize}

\section{Related Work}

\label{sec:related}


\noindent
\textbf{Egocentric VQA Benchmark}.\ There has been a growing interest in developing
benchmarks that systematically evaluate the spatial reasoning capabilities of multimodal
large language models (MLLMs)\cite{shiri2024empirical}. Most existing benchmarks~\cite{Lin2025OSTBenchET,Yang2025MMSIBenchAB,Azuma2021ScanQA3Q,Yang2024ThinkingIS}
formulate spatial reasoning VQA tasks using image sequences derived from 3D
scans or manually curated by researchers. Therefore, these works are largely
limited to object- or scene-centric geometric reasoning and overlook whether MLLMs
can understand 3D proximity in everyday human activities from a user-centric perspective,
as explored in our proposed EgoProx benchmark. A few egocentric VQA benchmarks have
been proposed to evaluate models' ability to reason about first-person behaviors~\cite{Zhou2025EgoTextVQATE,Mangalam2023EgoSchemaAD,Zhou2025EgoTextVQATE,Cheng2023EgoThinkEF}.
EgoSchema~\cite{Mangalam2023EgoSchemaAD} introduces visual question answer pairs
to test causality understanding of egocentric narratives. EgoPlan~\cite{Chen2023EgoPlanBenchBM}
focuses on goal-oriented reasoning from ongoing activities. Peng et al.~\cite{Peng2025InTE}
further developed a VQA benchmark that evaluates egocentric gaze-informed reason.
Huang et al.~\cite{Huang2025UnderstandingDS} introduced EgoDynamics4D, which
targets for 3D object- or agent-centric grounding. Although these works share a
similar motivation toward human-centric perception and behavior understanding, our
benchmark is the first to evaluate the cognitive reasoning of \textbf{3D
proximity} during daily activities.

\noindent
\textbf{Egocentric Multimodal Foundation Models.} MLLMs have achieved remarkable
progress in exocentric contexts~\cite{radford2021clip, li2022blip,
alayrac2022flamingo, chen2022pali, li2023blip2, liu2023llava, openai2023gpt4, bai2023qwenvl,
chen2024internvl, deepmind2024gemini, wang2024qwen2vl, wang2024qwen2vl}. Nevertheless, the
substantial domain gap between exocentric and egocentric visual--language data~\cite{lai2024lego}
greatly limits the generalization of exocentric-trained models when applied to first-person
scenarios. Recent advances in egocentric multimodal learning have introduced
specialized pretraining paradigms that explicitly align vision and language
representations from a first-person perspective. Egocentric captioning benefits from cross-view
or instructional adaptation strategies~\cite{xu2024retrieval, ohkawa2025exo2egodvc},
while question answering evolves toward temporally grounded reasoning~\cite{hou2023groundnlq,
wang2023lifelongmemory}. Kurita~\etal~\cite{kurita2023refego} and Sun~\etal~\cite{sun2025intentiongrounding}
extend language grounding to dynamic, interaction-rich scenes. Moreover,
language-guided action generation and instruction following~\cite{lai2024lego, min2021film}
connect egocentric observation with embodied decision-making. The pioneering EgoVLP~\cite{lin2022egovlp}
and EgoVLPv2~\cite{pramanick2023egovlpv2} conducted large-scale video--language
pretraining using Ego4D narrations. Zhao~\etal~\cite{zhao2023learning} and Suglia~\etal~\cite{suglia2024alanavlm}
refined visual--language alignment for long egocentric video understanding. More recently, Yang~\etal~\cite{yang2025egolife} proposed an omni-modal system that integrates models such as EgoGPT and EgoRAG to unify perception, reasoning, and interaction for egocentric understanding, while GazeGPT~\cite{konrad2024gazegpt} augmented large MMLMs with additional inputs to improve contextual reasoning for smart eyewear. Despite these advancements,
existing egocentric MLLMs remain limited in addressing 3D spatial reasoning, underscoring
the importance of our proposed EgoProx Benchmark.

\noindent\textbf{Spatial Intelligence}.\ Recent studies have suggested that existing multimodal large language models (MLLMs) still exhibit limitations in spatial intelligence\cite{chen2024spatialvlm,hong20233d},
motivating growing efforts to enhance this capability\cite{hong20233d,fu2024scene,chen2024spatialvlm,zheng2025video,wu2025spatial,daxberger2025mm}. Prevailing methods have attempted to directly encode 3D information such as point clouds\cite{chen2023ll3da,hong20233d,fu2024scene}, multi-view images\cite{hong20233d,zhu2025llava3d}
and objects\cite{chen2023ll3da,wang2023chat3d,huang2024leo,huang2024chatscene} as the context of MLLMs, following the footsteps of vision-language models (VLMs) to bridge the gap between 3D and language representations. Another line of work aims to enhance the spatial reasoning capabilities of MMLMs using only 2D inputs, such as images~\cite{daxberger2025mm,chen2024spatialvlm} or videos~\cite{wu2025spatial,Qi2024GPT4Scene}. Chen~\etal~\cite{chen2024spatialvlm} proposed SpatialVLM, which leverages well-developed 3D computer vision techniques such as monocular depth estimation, semantic segmentation, and region captioning to extract 3D spatial information from 2D images, using it as QA pairs to train MLLMs for spatial understanding. The Spatial-MLLM proposed by Wu~\etal~\cite{wu2025spatial} effectively encodes keyframes from videos into 3D information by incorporating the VGGT\cite{wang2025vggt} backbone, which is then fused with 2D embeddings before being input into MLLMs. Our benchmark also aims to evaluate the spatial reasoning capabilities of MLLMs without relying on explicit 3D representations as auxiliary modalities. This design choice aligns with the observation that humans can naturally infer approximate 3D proximity and spatial relationships from purely 2D visual inputs.


\section{EgoProx Benchmark}
\label{sec:benchmark}

In this section, we first introduce the formal definitions of the four task categories in our proposed EgoProx benchmark. We then describe the data sources and highlight the key features of the benchmark.

\subsection{Task Definition}
\label{sec:task_def}


We categorize proximity reasoning tasks along a cognitive hierarchy: human \emph{intention} shifts toward intermediate goals, driving both \emph{exploration} and \emph{exploitation} of the 3D environment. In addition, we include a more challenging \emph{chain-of-actions} reasoning task, which requires models to infer the multi-step proximity reasoning process underlying complex actions.

Formally, we define input video segments $\mathcal{X}=\left\{x_1,x_2,\ldots,x_T\right\}$, where $T$ is the total number of frames, $x_T$ denotes the current frame, and ${x_1, \ldots, x_{T-1}}$ represents the past frames. Each task examines the ability of the model $f_{\theta}$ to infer the correct answer $\mathcal{A}$ from a discrete set of candidates $\mathcal{C}$, given a natural language question $\mathcal{Q}$ and $\mathcal{X}$.

\noindent\textbf{Exploration} evaluates whether $f_{\theta}$ can predict the navigation step $\hat{s}$ toward the goal $G$, with the goal specified by the query $Q$ and visible within the input video segment $\mathcal{X}$.

\noindent\textbf{Exploitation} assesses whether $f_{\theta}$ can predict how the next human-object interaction $\hat{h}$ will happen in 3D space, given the observable segment $\mathcal{X}$ and the query $Q$ describing the ongoing manipulation context.

\noindent\textbf{Intention} examines whether $f_{\theta}$ can predict immediate body movements $\hat{m}$, including gaze shifts or head movements conditioned on the goal $G$ specified by the query $Q$, based on the observable segment $\mathcal{X}$.

\noindent\textbf{Chain-of-Actions Reasoning} assesses whether $f_{\theta}$ can predict a sequence of future actions $\{a_1, a_2, \ldots, a_K\}$ and their relative spatial relationships $\{e_i\}$ of action locations, given the observable segment $\mathcal{X}$, a high-level goal $G$, and the query $Q$. Specifically, each $e_i$ encodes the spatial relation between consecutive locations $(l_i, l_{i+1})$, using the image plane of frame $x_T$ as the reference coordinate. We also provide number of steps $k$ and a candidate action set $\mathcal{S}$ consisting of the future actions ${a_1, \ldots, a_k}$ along with a set of distraction actions to limit the exploration space of MLLMs.

When designing the correct answer $\mathcal{A}$ and the distractor options, we consider two types of proximity measurements:
(1) \emph{Approximate proximity}, which encodes coarse metric transformations required at the last observable time step $T$, parameterized by angular rotations and translational displacements; and
(2) \emph{Relative proximity}, which represents discrete spatial relationships between a reference and a target entity at time $T$, characterized by  spatial predicates (e.g., left–right, front–back, near–far) that describe directional topology rather than absolute metric distance. Considering the challenging nature of the Chain-of-Actions task, we evaluate only the relative proximity for this task.

\subsection{Data Source}
\label{sec:source}
To construct our benchmark, we leverage the existing EgoExo4D~\cite{Grauman2023EgoExo4DUS} and Aria Digital Twin (ADT)~\cite{pan2023aria} datasets. Both capture egocentric video streams from fisheye cameras, along with calibrated poses and eye-tracking data. EgoExo4D provides upper-body pose annotations and atomic action descriptions but lacks 3D object annotations, whereas ADT offers dense 3D object annotations but lacks semantic action labels. Activities in EgoExo4D often occur in constrained environments with limited locomotion, making it unsuitable for exploration-related reasoning. In contrast, activities in ADT mainly involve walking within the scene and manipulating objects but lack goal-oriented behaviors, making it inadequate for chain-of-actions reasoning. The detailed distribution of benchmark data sources is provided in the Supplementary Materials.


\subsection{Benchmark Characteristics}

\begin{table}[t]
\caption{Comparison of EgoProx with existing 3D reasoning VQA or egocentric activity VQA benchmarks. We summarize key properties including 3D awareness, dataset scale, reasoning types, construction methodology, and temporal reasoning range. The reasoning types include grounding (G), forecasting (F), planning (P), and causality (C). Benchmark construction types include human annotation, MLLM/LLM-based generation, and agent-based generation. For clarity, note that human review for quality assurance is adopted by all existing QA-generation pipelines, including ours.}
\vspace{-1em}
  \resizebox{\columnwidth}{!}{

  \begin{tabular}{@{}c l c c l c c@{}}
    \toprule                                                    & \textbf{Benchmark}                           & \textbf{3D Space} & \textbf{$\#$ of Samples} & \textbf{Reasoning Type} & \textbf{Construction} & \textbf{Temporal Range} \\
    \midrule \multirow{6}{*}{\rotatebox{90}{No Action}}          %
                                                                & ScanQA\cite{Azuma2021ScanQA3Q}               & \cmark            & 46313                    & Grounding               & Human                      & Short                   \\ 
                                                                & MMSI-Bench\cite{Yang2025MMSIBenchAB}         & \cmark            & 1000                     & Grounding               & Human                      & Short                   \\ 
                                                                & VSI-Bench\cite{Yang2024ThinkingIS}           & \cmark            & 5000+                    & Planning                & Human                      & Short\& Long            \\ 
                                                                & OST-Bench\cite{Lin2025OSTBenchET}            & \cmark            & 10k                      & Grounding               & Human                      & Short                   \\ 
                                                                & OpenEQA\cite{majumdar2023openeqa}            & \cmark            & 1600+                    & Grounding               & Human                      & Short                   \\ 
                                                                & VLM4D\cite{zhou2025vlm4d}                    & \cmark            & 1800+                    & Grounding               & Human                      & Short                   \\ 
    \midrule \multirow{16}{*}{\rotatebox{90}{Egocentric Action}}
                                                                & EgoVQA\cite{fan2019egovqa}                   & \xmark            & 581                      & Grounding               & Human                      & Long                    \\ 
                                                                & EgoTextVQA\cite{Zhou2025EgoTextVQATE}        & \xmark            & 7064                     & Grounding               & MLLM                       & Short\& Long            \\ 
                                                                & EgoMemoria\cite{ye2025mmego}                 & \xmark            & 7026                    & Grounding               & LLM                        & Short\& Long            \\ 
                                                                & QAEgo4D\cite{Barmann2022ego4dqa}             & \xmark            & 1854                     & Grounding               & Human                      & Short\& Long            \\ 
                                                                & AssistQ\cite{benita2022assistq}              & \xmark            & 531                      & Grounding               & Human                      & Long                    \\ 
                                                                & Ego-ST\cite{wu2025stthink}                   & \cmark            & 5000+                    & G\&P     & Human                      & Long                    \\ 
                                                                & EgoDynamics4D\cite{Huang2025UnderstandingDS} & \cmark            & 927K                     & Grounding               & MLLM                       & Short\& Long            \\ 
                                                                & EgoThink\cite{Cheng2023EgoThinkEF}           & \xmark            & 700                      & G\&F\&P                   & Human                      & Short                   \\ 
                                                                & EgoTaskQA\cite{jia2022egotaskqa}             & \xmark            & 40K                      & Grounding               & LLM                        & Short                   \\ 
                                                                & EOC-Bench\cite{yuan2025eoc}             & \xmark            & 3277                      & G\&F                & Human                        & Short                   \\ 
                                                                & VideoMindPalace\cite{huang2025building}      & \xmark            & 1757                    & Grounding               & MLLM                       & Short\& Long            \\ 
                                                                & EgoGazeVQA\cite{Peng2025InTE}                & \xmark            & 1800                     & G\&C    & MLLM                       & Short                   \\ 
                                                                & EgoSchema\cite{Mangalam2023EgoSchemaAD}      & \xmark            & 5063                     & Causality               & LLM                        & Long                    \\ 
                                                                & EgoPlan\cite{Chen2023EgoPlanBenchBM}         & \xmark            & 4939                    & Planning                & MLLM                       & Long                    \\ 
                                                                & EgoLifeQA\cite{yang2025egolife}            & \xmark            & 6000                     & Grounding               & LLM                        & Long                    \\ 
    \rowcolor{mygray}                                           & \textbf{EgoProx (Ours)}                      & \cmark            & 2405                     & G\&F\&P                   & Agent                      & Short\& Long            \\ 
    \bottomrule
  \end{tabular}
  }

  \label{tab:benchmak_charac}
\end{table}

As shown in Table \ref{tab:benchmak_charac}, we provide a detailed comparison between our EgoProx and related benchmarks in egocentric vision and spatial intelligence. Our EgoProx encompasses a broad spectrum of reasoning tasks and represents the first benchmark to assess 3D spatial intelligence in the context of human behavior. While most existing VQA benchmarks categorize tasks by reasoning type (e.g., grounding, planning, forecasting), we instead characterize our dataset according to the cognitive hierarchy as introduced earlier. This design choice stems from the coupled nature of intention perception and action execution in egocentric videos. It is worth noting that the question types in our benchmark still resonate with prior efforts. As detailed in Sec.~\ref{sec:task_def}, the Intention, Exploration, and Chain of Actions categories evaluate a model’s ability to infer subsequent steps based on the current state, thereby aligning closely with grounding and planning tasks. In contrast, the Exploitation category assesses the model’s ability to predict short-horizon or intermediate future events, corresponding to the forecasting dimension defined in existing benchmarks. Notably, such a comprehensive benchmark requires a carefully designed data construction pipeline, even when leveraging datasets like EgoExo4D and ADT that already include valuable annotations and modalities.

\begin{figure*}
    \centering
    \includegraphics[width=1\linewidth, keepaspectratio]{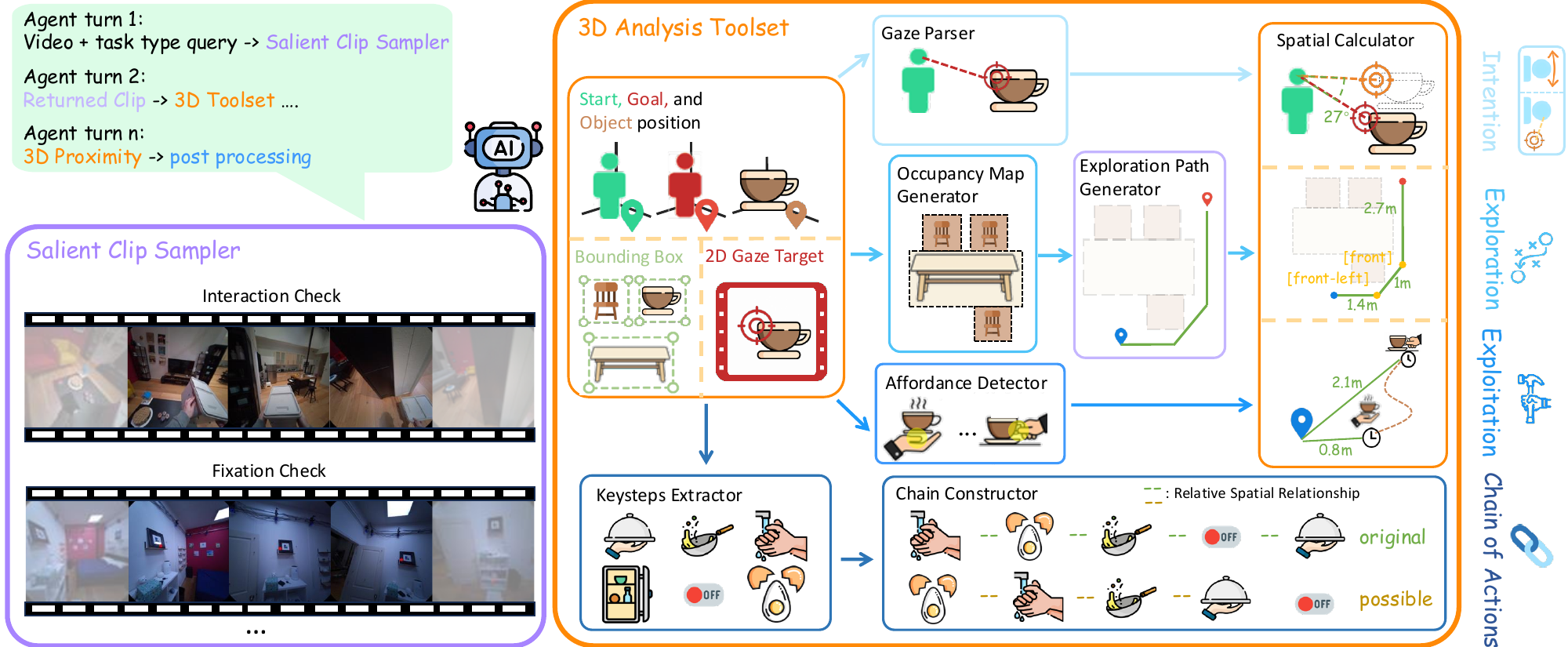}
    \caption{\textbf{Overview of our agent-based data construction pipeline}.The agent first identifies salient moments with an interaction- and fixation-based sampler, then uses the 3D Analysis Toolset to extract spatial cues such as object positions, gaze targets, occupancy maps, and action chains. It then invokes the Spatial Calculator to derive 3D distances, orientations, and proximity relations, producing structured 3D proximity ground truth. Final benchmark question-answer pairs are compiled through necessary post-processing.}
    \label{fig:tool}
\end{figure*}
\section{Agentic Data Engine}
\label{sec:engine}

As shown in~\ref{fig:tool}, given an egocentric video, along with its associated metadata (camera pose, objects and action labels etc.), and a user-specified question category, our agent uses Gemini-2.5-Pro as the foundation and orchestrates a suite of specialized tools to synthesize question–answer pairs in a controllable manner. We introduce the key components of our agent in the following sections.



\subsection{Salient Clip Sampler}

The first step of our data engine is to extract an ideal clip $\mathcal{X}=\{x_1,\ldots,x_T\}$ from a long egocentric video. 
Although datasets such as EgoExo4D provide coarse temporal annotations, our tasks require finer alignment to ensure that each question is answerable from the last observable frame $x_T$. 
We therefore adopt a unified sampling principle that organizes all tasks into two functional categories. 
Forecasting tasks, such as gaze forecasting in \emph{Intention} and next-interaction prediction in \emph{Exploitation}, are defined by future supervisory events. 
We detect the moment where a stable gaze fixation or a human--object interaction occurs, and then select the video segment preceding this event so that $\{x_1,\ldots,x_T\}$ implicitly encodes the preparatory cues leading into the upcoming behavior. 
Planning tasks, including head-orientation reasoning in \emph{Intention}, \emph{Exploration}, and \emph{Chain of Actions} tasks, require clips that provide partial but sufficient evidence for achieving a given goal $G$. 
For Exploration and head-orientation reasoning, we enforce that $G$ is visible in some earlier frame $x_t$ but not in the final observation $x_T$, determined through field-of-view visibility checks. 
For Chain of Actions, we identify dense regions of keysteps, derive a high-level goal from the earlier portion of the window, and ensure that several future steps remain after $x_T$ so that the multi-step plan is still inferable. 
This task-driven clip sampler guarantees that each extracted video segment contains the minimal yet sufficient cues required for the intended form of 3D proximity reasoning.

\subsection{Toolset for 3D Analysis}

In Fig. 2, we provide a visual illustration of our toolsets and how they are used to construct the ground truth for the four task categories. We briefly introduce the input and output of each tool here, with implementation details provided in the Supplementary Materials.

\noindent\textbf{Occupancy Map Generator} constructs an occupancy map from the 3D bounding boxes associated with $\mathcal{X}$, identifying free or occupied regions for obstacle checking. 

\noindent\textbf{Exploration Path Generator} computes a feasible path within the occupancy map from the query position to the goal position using an 8-connected A$^*$ search algorithm. Note that for navigation step generation, we do not use the actual camera trajectory to derive this path, as human motion is inherently stochastic and poses significant challenges for MLLMs to interpret reliably.


\noindent\textbf{Spatial Calculator} consists of the Distance Calculator and the Direction Calculator. 
The Distance Calculator projects both camera and object positions into a unified world coordinate system and computes the translation distances between the queried entities.
While the Direction Calculator returns the angle between the camera’s forward direction and the vector from the camera position to the target $G$, projected onto the Bird’s-Eye View (BEV).

\noindent\textbf{Gaze Parser} transforms eye-tracking data from the 2D image plane into a corresponding 3D gaze ray, which is then used to localize the object being fixated.


\noindent\textbf{Affordance Detector} determines whether the target object $G$ is interacted with in the future frames, and further computes, in the current frame $\mathcal{X}_T$, the direction and distance from the observer to $G$ using aforementioned direction and distance calculators.


\noindent\textbf{Keystep Extraction Tool} returns the textual keysteps including the interactive objects, the observer, and the interaction names in the observation video.

\noindent\textbf{Chain Constructor} obtains possible chains of steps and the direction between the steps. First, the chain tool calculate the directions between the steps. Regarding it as the basically correct chain, the tool provides several possible correct chains using multi-modal large language models.

\subsection{Toolset Usage}
In a nutshell, the 3D proximity ground truth for a given input clip sampled for each task type is constructed as follows:
\begin{itemize}
    \item \textbf{Intention}: The agent invokes the Spatial Calculator to estimate how the camera wearer adjusts head orientation toward the goal or directs gaze inferred by the Gaze Parser.
    \item \textbf{Exploration}: The agent samples a valid goal $G$ based on visibility checks and adopts the Occupancy Map Generator and Exploration Path Generator to obtain a path composed of steps $\hat{s}$, each providing the distance and discrete direction for exploration.
    \item \textbf{Exploitation}: 
    The agent utilizes an affordance detector to identify 
    which part of the object $G$ the observer is grasping in the anticipation frame, 
    where the observer will place the object $G$, 
    and which direction the observer will move to interact with the object $G$.
    \item \textbf{Chain of Actions}: Specifically, the agent employs the Keystep Extractor to extract key action steps and their 3D spatial locations from long video segments, and to identify the key actions toward the common goal $G$ based on future observations. It then employs an LLM to construct a set of all possible ordered combinations of key steps leading toward the same goal. Finally, The agent calls the Chain Constructor to generate a complete set of possible answers by calculating the spatial relationships among the ordered combinations of key steps.
\end{itemize}

\subsection{Post-Processing}

As mentioned in Sec.~\ref{sec:task_def}, the proximity measurements include both approximate transformation and relative relationships. We discretize the transformation into intervals that are interpretable by humans. For spatial relationships, we convert the 3D directions into eight discrete orientations projected onto a specified plane. When constructing the candidate sets, we prompt the VLM to generate hard-negative distraction options. However, we provide specific instructions to ensure that these distractions do not rely on minor differences that are unsolvable even for humans.

We also conduct careful human verification to ensure both the answerability and accuracy of the ground truth. For the \emph{Chain of Actions} task, we perform a thorough examination of all possible answer sets generated by the agent.

\begin{table*}[htbp]
\small
\centering
\setlength{\tabcolsep}{4pt}

\caption{Evaluation results of prevailing MLLMs on the EgoProx benchmark, where best scores are colored with \colorbox{red!30}{red} and the second best scores are colored with \colorbox{orange!40}{orange}. All models are evaluated using a unified prompt that defines the egocentric, world, and image-plane coordinate systems, and adopts zero-shot chain-of-thought prompting following~\cite{wei2022chain}.}
\vspace{-1em}
\begin{tabular}{l*{10}{c}}
\toprule
\multirow{2}{*}{Model} 
& \multicolumn{2}{c}{Intention} 
& \multicolumn{2}{c}{Exploration} 
& \multicolumn{2}{c}{Exploitation} 
& \multicolumn{3}{c}{Chain of Actions} \\
\cmidrule(lr){2-3}
\cmidrule(lr){4-5}
\cmidrule(lr){6-7}
\cmidrule(lr){8-10}
& Approx. & Relative 
& Approx. & Relative 
& Approx. & Relative 
& \emph{Act-Acc} & \emph{Rel-Acc-S} & \emph{Rel-Acc-L} \\
\midrule

Human Level & 62.50 & 75.33 & 60.00 & 63.15 & 82.02 & 85.25 & 80.23 & 63.25 & 83.12 \\

\midrule

\multicolumn{10}{l}{\textit{Proprietary Models}} \\
Gemini-2.5-Pro   & \cellcolor{red!30}42.75 & \cellcolor{orange!40}37.13 & \cellcolor{orange!40}36.90 & \cellcolor{orange!40}29.32 & \cellcolor{red!30}50.24 & \cellcolor{red!30}45.17 & \cellcolor{red!30}25.14 & 17.03 & \cellcolor{orange!40}52.36 \\
GPT-5            & 33.16 & \cellcolor{red!30}40.35 & \cellcolor{red!30}41.18 & \cellcolor{red!30}34.55 & 46.45 & \cellcolor{red!30}45.17 & \cellcolor{orange!40}21.74 & \cellcolor{orange!40}20.83 & \cellcolor{red!30}52.71 \\ 
\midrule

\multicolumn{10}{l}{\textit{Open-source Models}} \\
LLaVA-NeXT-Video-7B & 23.06 & 27.19 & 18.72 & 23.56 & 31.04 & 29.29 & 1.09 & 16.67 & 33.33 \\
MiniCPM-V 2.6        & 28.50 & 29.24 & 22.99 & 9.42 & 37.44 & 31.60 & 2.63 & \cellcolor{red!30}25.00 & 50.00 \\
InternVL 2.5-8B      & 26.94 & 28.95 & 18.72 & 19.37 & 36.02 & 33.77 & 8.25 & 6.25 & 52.08 \\
Qwen2.5-VL-7B        & 33.68 & 29.24 & 27.27 & 20.42 & 38.63 & 34.34 & 5.98 & 2.27 & 46.21 \\
Qwen2.5-VL-32B       & 31.35 & 33.92 & 30.48 & 17.80 & 45.97 & 40.55 & 10.33 & 7.02 & 45.61 \\
Qwen2.5-VL-72B       & 30.83 & 35.38 & 29.41 & 24.08 & 46.21 & 40.26 & 13.04 & 14.24 & 48.61 \\
Qwen3-VL-235B        & \cellcolor{orange!40}34.46 & 33.33 & 28.34 & 27.75 & 45.97 & 42.28 & 10.87 & 11.67 & 51.25 \\
Qwen3-VL-Plus        & 26.42 & 34.21 & 20.32 & 21.47 & \cellcolor{orange!40}48.34 & 42.28 & 10.87 & 7.50 & 41.67 \\
\bottomrule
\end{tabular}
\label{tab:benchmark}
\end{table*}

\begin{table*}[htbp]
\small
\centering
\setlength{\tabcolsep}{4pt}

\caption{Cross-category experimental results where best scores are colored with \colorbox{red!30}{red}. We leverage extra training data from one category generated by our data engine and evaluate performance across all categories. The additional data not only improves performance within the source category but also enhances cross-category generalization, revealing the inherent hierarchical structure of human cognition.}
\vspace{-1em}
\begin{tabular}{l*{10}{c}}
\toprule
\multirow{2}{*}{Model} & \multicolumn{2}{c}{Intention} & \multicolumn{2}{c}{Exploration} & \multicolumn{2}{c}{Exploitation} & \multicolumn{3}{c}{Chain of Actions} \\
\cmidrule(lr){2-3}
\cmidrule(lr){4-5}
\cmidrule(lr){6-7}
\cmidrule(lr){8-10}
& Approx. & Relative 
& Approx. & Relative 
& Approx. & Relative 
& \emph{Act-Acc} & \emph{Rel-Acc-S} & \emph{Rel-Acc-L} \\
\midrule
Qwen2.5-VL-7B & 33.68 & 29.24 & 27.27 & 20.42 & 38.63 & 34.34 & 5.98 & 2.27 & \cellcolor{red!30}46.21\\
\midrule
Qwen2.5-VL-7B + Intention Tuning & -- & -- &  \cellcolor{red!30}32.09 &\cellcolor{red!30} 24.61 & \cellcolor{red!30}64.93 & \cellcolor{red!30}39.11 & 3.80 & 0.00 & 14.29 \\
Qwen2.5-VL-7B + Exploration Tuning & 45.34 & 35.09 & -- & -- & 45.26 & 34.34 & \cellcolor{red!30}7.61 & \cellcolor{red!30}14.29 & 40.48 \\
Qwen2.5-VL-7B + Exploitation Tuning & \cellcolor{red!30}56.48 & \cellcolor{red!30}36.55 & 27.27 & 20.42 & -- & -- & 4.35 & 0.00 & 16.67 \\
\bottomrule
\end{tabular}
\label{tab:task_result}
\end{table*}

\begin{table}[t]
\centering
\small
\setlength{\tabcolsep}{3pt}

\caption{Cross-dataset experimental results. Fine-tuning on one dataset improves proximity reasoning on the other.}
\vspace{-1em}
\begin{tabular}{lcccc}
\toprule
\multirow{2}{*}{Model} 
& \multicolumn{3}{c}{ADT Evaluation} \\
\cmidrule(lr){2-4}
& Intention & Exploration & Exploitation \\
\midrule
Qwen2.5-VL-7B              & 35.94 & 23.81 & 47.64 \\
EgoExo4D Only Tuning & 48.70 & 27.78 & 64.57 \\
\midrule
\multirow{2}{*}{Model} 
& \multicolumn{3}{c}{EgoExo4D Evaluation} \\
\cmidrule(lr){2-4}
& Intention & Exploration & Exploitation \\
\midrule
Qwen2.5-VL-7B              & 26.74 & --    & 32.52 \\
ADT Only Tuning   & 50.58 & --    & 43.67 \\
\bottomrule
\end{tabular}
\label{tab:data_result}
\end{table}

\section{Experiments}
\label{sec:Experiments}

\subsection{Metrics}
\label{sec:metrics}

The majority of our benchmark consists of multiple-choice questions; therefore, we adopt a straightforward accuracy metric with a 20\% chance level. 

For the Chain of Action reasoning evaluation, we explicitly structure model outputs as a chain of nodes defined in~\ref{sec:task_def}, where each node represents a selected action step and the edge values encode their relative spatial relationships. In our benchmark, each sample consists of 3–5 steps, and the candidate action set $\mathcal{S}$ contains 10 candidates. And our agent generates a ground-truth answer set $\mathcal{Y}$ that encompasses all valid possibilities, and the size of the valid ground-truth set $\mathcal{Y}$ ranges from 1 to 3. The Action Accuracy \emph{(Act-Acc)} is computed by comparing the predicted ordered nodes ${o_1, \ldots, o_k}$ with those in the ground-truth set. 

For correctly predicted sequences, we further evaluate the spatial relationship accuracy, denoted as Relational Accuracy \emph{(Rel-Acc-S)}, defined as $c / (k - 1)$, where $c$ is the number of correctly predicted relationships and $(k - 1)$ is the total number of edges. To account for ambiguity in action locations, we also introduce a relaxed version, Relational Accuracy–Loose \emph{(Rel-Acc-L)}, where a predicted orientation (e.g., front-right) is considered correct if the ground truth belongs to one of its adjacent directions (e.g., front, right, or front-right).

\subsection{Results on EgoProx Benchmark}

We first evaluate prevailing proprietary API-based models, including GPT-5~\cite{openai25-gpt} and Gemini-2.5-Pro~\cite{deepmind2024gemini}, as well as several recent open-source models, such as LLaVA-NEXT-Video-7B~\cite{li2024llava}, MiniCPM-V 2.6~\cite{yao2024minicpm}, InterVL 2.5~\cite{chen2024expanding}, and the Qwen-VL series~\cite{Qwen2.5-VL} across different model scales. For all models, we use a unified inference prompt to ensure a fair comparison. The prompt specifies the task and response constraints, includes the question text, and provides a minimal output-format exemplar to enable deterministic parsing. We also employ a zero-shot reasoning-style prefix~\cite{wei2022chain} that encourages step-by-step inference. We provide the exact prompt template in the Supplementary Materials.

We provide detailed experimental results in Tab.~\ref{tab:benchmark}. Consistent with previous finds~\cite{Yang2025MMSIBenchAB}, even the most advanced proprietary models still struggle with 3D proximity reasoning compared to human-level capability. Particularly, humans perform consistently well on the Chain of Actions task, but MLLMs drop sharply relative to other tasks, highlighting the difficulty of long-horizon reasoning. Proprietary models slightly outperform their open-source counterparts, particularly on exploration tasks, likely due to large-scale pretraining corpora that include long video sequences demonstrating how agents traverse complex environments. In addition, the Qwen-VL series achieves the strongest overall performance among open-source models. However, unlike general VQA benchmarks, scaling up model size yields only limited performance gains, a trend consistent with recent findings in 3D spatial understanding VQA benchmarks~\cite{Yang2025MMSIBenchAB,Lin2025OSTBenchET}.

In this context, we pose a critical question: does the limited performance of existing models indicate an inherent absence of spatial intelligence, or does it instead reflect their inability to utilize the spatial knowledge implicitly encoded within their large-scale parameters when addressing spatial reasoning queries? In the following section, we conduct additional experiments to further investigate this question.

\subsection{Additional Analysis}
\label{subsec:additional_analysis}

\noindent\textbf{Hypothesis}.\ Existing MLLMs should have gained latent spatial knowledge during pretraining, as the massive multimodal data consisting of image–text pairs, video captions, and related sources contain abundant implicit cues about geometry, spatial relationships, and affordances. However, this knowledge is often entangled and implicitly represented, making it difficult to retrieve for structured reasoning tasks, which leads to suboptimal performance on various spatial AI benchmarks, including ours. 

\noindent\textbf{Experiment Setup}.\
We first utilize the aforementioned data engine to construct additional instruction-tuning data that has no overlap with the testing set, ensuring a fair evaluation. We then LoRA fine-tune Qwen2.5-VL-7B using the LLaMA-Factory framework~\cite{zheng2024llamafactory} on a small set of training data from one single data source or task category, and then evaluate the model’s cross-data and cross-task performance. Note that we use 800 samples for cross-task experiments and 1,200 samples per category for cross–dataset experiments. Given its limited scale, this training data is unlikely to introduce new knowledge into the MLLMs, specifically with visual encoder frozen. Instead, it primarily aims to guide the models in better utilizing the spatial knowledge already embedded within their parameters. We provide detailed training recipe in the supplementary materials.

\begin{figure*}
    \centering
    \includegraphics[width=1\linewidth, keepaspectratio]{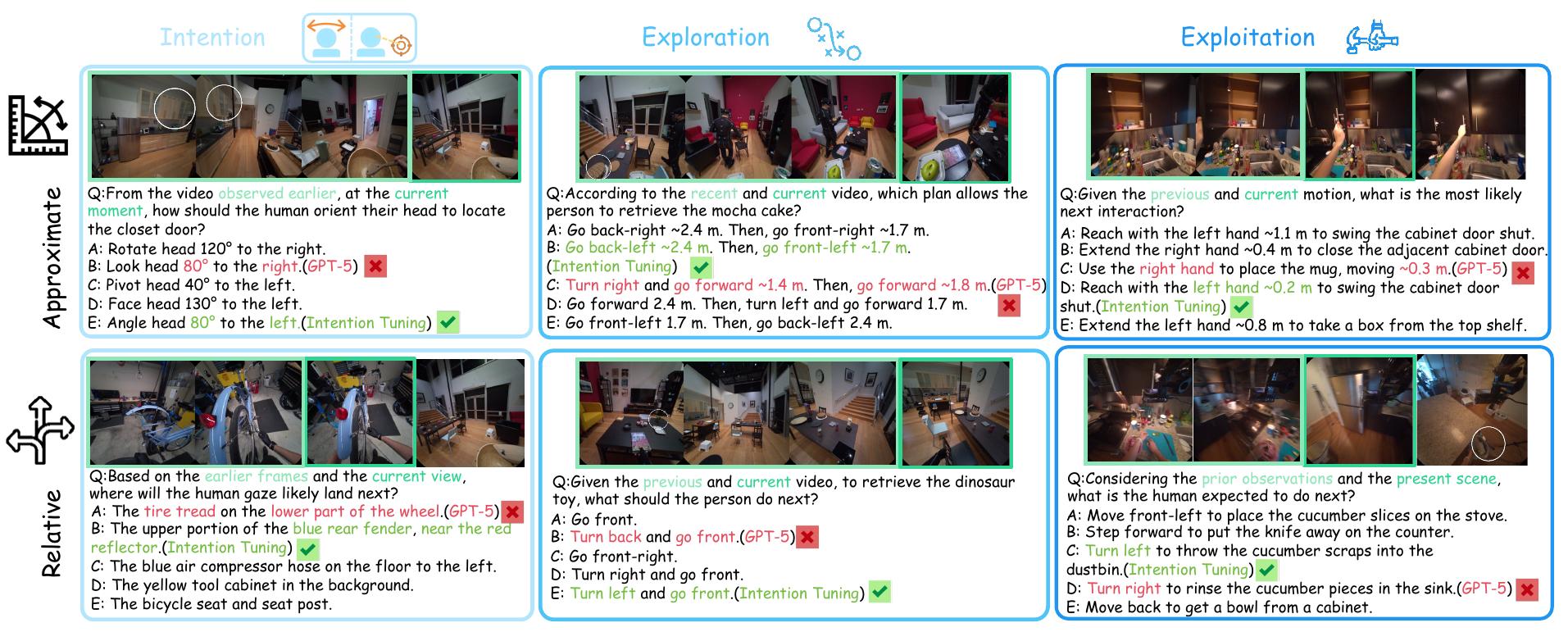}
    \vspace{-2em}
    \caption{\textbf{Visual examples of our benchmark and model performance.} We show cases where the intention-tuned model outperforms the proprietary GPT-5 model. }
    \label{fig:qa}
\end{figure*}

\noindent\textbf{Task-Specific Instructing Tuning}.\ Tab.~\ref{tab:task_result} presents the results of the cross-category instruction tuning experiments. Notably, using a small amount of training data from one task often leads to improvements on other tasks. This provides strong evidence that the model already possesses latent spatial knowledge, but cannot effectively leverage it through zero-shot prompting alone. Another interesting observation is that although all three fine-tuning experiments use the same amount of data, the Intention training data yields a notably larger performance gain on other tasks compared to the Exploration or Exploitation training data. This aligns with our key motivation for organizing the benchmark along a cognitive hierarchy: intention provides the fundamental signals that guide both location and action. Concretely, understanding intentional cues is key to driving action-conditioned 3D reasoning, thereby providing additional insight into the instruction tuning of human-centric, 3D-aware MLLMs.

We further report model performance on the Chain-of-Actions task under task-specific tuning in Tab.~\ref{tab:task_result}. Both Intention tuning and Exploitation tuning lead to a slight decrease in model performance, as these data do not contain multi-step reasoning signals. In contrast, Exploration tuning, although focused primarily on navigation steps, still provides useful supervision on action locations and therefore improves multi-step reasoning. These results further confirm our hypothesis that existing MLLMs contain latent spatial intelligence, yet depend on instruction tuning to effectively express and utilize this capability.

\noindent\textbf{Dataset-Specific Instructing Tuning}.\ We conduct similar experiments under a cross-dataset setting. As shown in Tab.~\ref{tab:data_result}, we observe substantial performance improvements despite the large recording domain gap between ADT and EgoExo4D. Note that the improvement on the ADT Exploration task is smaller, primarily because the EgoExo4D training data do not contain any Exploration-type questions, as explained in Sec.~\ref{sec:source}.

\subsection{Visual Illustrations}
 
We provide visualization of the performance of GPT-5 and the instruction-tuned Qwen2.5-VL-7B(using only intention-type data) on our benchmark in Fig.~\ref{fig:qa}.

GPT-5 often produces answers that appear semantically reasonable, yet it struggles with spatial reasoning, frequently failing to correctly interpret egocentric relative positions and spatial relationships. Moreover, it is unable to reliably connect cues such as spatial evidence and intention signals from the observed video to the actions that are about to occur.In contrast, the intention-tuned model consistently aligns its forecasting or planning with cues from the past video, yielding correct results. This observation also aligns with the cognitive hierarchy that motivates our benchmark design, where intention informs locomotion and reaching, and ultimately supports hierarchical interactions in complex 3D scenes. Additional visualizations and failure-case analyses are provided in the supplementary material.









\section{Conclusion}

In this paper, we present \textbf{EgoProx}, the first benchmark for egocentric 3D proximity reasoning. The benchmark is organized as a cognitive hierarchy with four tasks, progressing from Intention to Exploration, Exploitation, and Chain of Actions. We further introduce an agent-based data engine with a suite of tools that enables scalable and high-quality data generation. Extensive experiments reveal key spatial reasoning bottlenecks in current MLLMs. Cross-domain instruction tuning results suggest that the limited spatial understanding of MLLMs arises not from missing spatial knowledge,
but from ineffective mechanisms for leveraging knowledge already encoded in model parameters.




\noindent\textbf{Acknowledgments}.\ This work was supported in part by the Zhiyuan Scholar Program from the Beijing Municipal Science and Technology Commission (Z251100008125045) and NSFC Grants.

{
    \small
    \bibliographystyle{ieeenat_fullname}
    \bibliography{main}
}

\clearpage
\maketitlesupplementary

\renewcommand\thesection{\Alph {section}}
\renewcommand\thesubsection{\thesection.\arabic{subsection}}

\setcounter{section}{0}

This is the supplementary material for the paper “EgoProx: Evaluating MLLMs on Egocentric 3D Proximity Reasoning Across a Cognitive Hierarchy". We organize the content as follows.
\\

\noindent\textbf{\hyperref[sec:evaluation]{A} --  Evaluation Details} \\[0.2cm]
\noindent\textbf{\hyperref[sec:statistics]{B} -- Benchmark Statistics} \\ [0.2cm]
\noindent\textbf{\hyperref[sec:tool]{C} -- Implementation Details of Toolset} \\ [0.2cm]
\noindent\textbf{\hyperref[sec:exp]{D} -- Additional Analysis on the experimental Results} \\ [0.2cm]
\noindent\textbf{\hyperref[sec:train]{E} -- Training Details on Domain-specific Tuning} \\ [0.2cm]
\noindent\textbf{\hyperref[sec:vis]{F} -- Additional Visualization} \\ [0.2cm]
\noindent\textbf{\hyperref[sec:limitation]{G} -- Limitations} \\ [0.2cm]
\noindent\textbf{\hyperref[sec:prompt]{H} -- Prompt Template for Evaluation} \\ [0.2cm]
\noindent\textbf{\hyperref[sec:train_prompt]{I} -- Prompt Template for Training}

\section{Evaluation Details}
\label{sec:evaluation}

\noindent\textbf{General Evaluation Setup}.\
For all evaluation processes conducted on our benchmark, we first uniformly sample each video into 8 frames. To ensure reproducibility, unless otherwise specified, we adopt a greedy decoding strategy for all models (\textit{i.e.}, the temperature is set to $0$, and both top-p and top-k are set to $1$). The multimodal input to each model is formatted as follows: \textit{[video frames] [text prompt]}. We use a unified inference prompt to ensure a fair comparison across models. The text prompt specifies the task objective and response constraints, incorporates the question text, and includes a minimal output-format exemplar to facilitate deterministic parsing during evaluation. Additionally, we append a zero-shot reasoning prefix~\cite{wei2022chain} to encourage step-by-step inference behaviors commonly observed in instruction-tuned MLLMs. The exact prompt templates used for each task category are detailed in Section~\ref{sec:prompt}.

\noindent\textbf{Human Level Performance}.\
To assess human-level performance on \textbf{EgoProx}, we adopt an evaluation procedure inspired by prior benchmarking protocols such as VSI-Bench~\cite{Yang2024ThinkingIS}. Human participants receive both the question and its corresponding video sequence simultaneously and are allowed unlimited time to provide their responses. To conduct the evaluation, we sample a representative subset of our benchmark, selecting 50 questions per task category to ensure balanced task coverage. We recruit individuals who possess basic familiarity with spatial AI and MLLMs, and we supply clear instructions along with illustrative examples. Participants may replay the video as many times as needed to ensure thorough understanding of video context before making a decision.

\section{Benchmark Statistics}
\label{sec:statistics}
\begin{figure}[t]
    \centering
    \includegraphics[width=\linewidth]{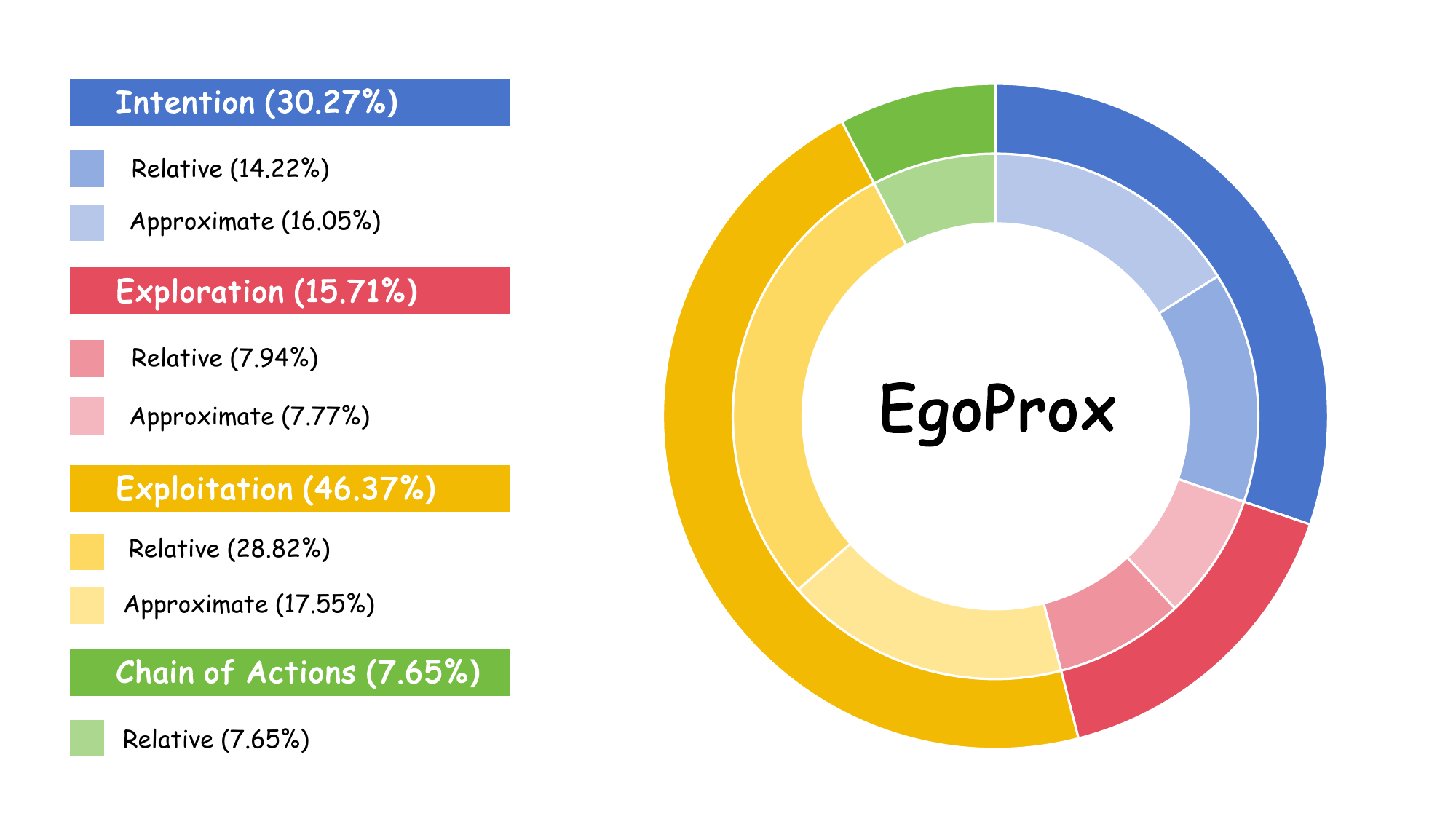}
    \vspace{-2em}
    \caption{\textbf{Benchmark Statistics.} The distribution of tasks across four main categories in \textbf{EgoProx} with Relative and Approximate variants. }
    \label{fig:chart}
\end{figure}

EgoProx contains 2,405 VQA samples, encompassing a broad spectrum of egocentric 3D proximity reasoning tasks.
These samples are derived from two complementary egocentric datasets: 1,016 from Aria Digital Twin (ADT)\cite{pan2023aria} and 1,389 from EgoExo4D\cite{Grauman2023EgoExo4DUS}. Due to differences in dataset characteristics, task coverage varies across sources: \textit{Exploration} tasks are exclusively generated from ADT, where locomotion is prominent, whereas \textit{Chain of Actions} tasks rely solely on EgoExo4D, which contains dense, goal-oriented manipulation sequences. For the remaining task categories, samples are drawn from both datasets with balanced proportions.

As shown in Fig.~\ref{fig:chart}, the benchmark is structured across four primary categories: \textit{Intention (30.27\%)}, \textit{Exploration (15.71\%)}, \textit{Exploitation (46.37\%)} and \textit{Chain of Actions (7.65\%)}, reflecting the cognitive hierarchy introduced in the main paper. Except for \textit{Chain of Actions}, each task category includes two distinct forms of proximity measurement: \textit{Relative} and \textit{Approximate}.

\section{Implementation details of Toolset}
\label{sec:tool}

\begin{table}
    
    \small
    \renewcommand{\arraystretch}{1.5}
        \caption{\textbf{Summary of input notation.} For simplicity, we omit the time step for some of the notations.}
        \vspace{-0.8em}
        \resizebox{\columnwidth}{!}{\begin{tabular}{cc}
            \toprule
            Notation & Definition \\
            \midrule
            $\mathcal{X}=\left\{x_1,x_2,\ldots,x_T\right\}$ & Observable video segment, where $T$ denotes the total number for frames\\
            $\mathcal{O}^{3d}=\left\{o_1^{3d},o_2^{3d},...,o_N^{3d}\right\}$ & 3D object bounding boxes, where N denotes the total number of objects\\
            $E$ & Eye gaze data including the pose and the depth \\
            $S$ & Skeleton position data, including hand skeleton position\\
            $T_{s}^{d}$ & Transformation matrix between scene and device\\
            $T_{c}^{d}$ & Transformation matrix between device and camera\\ 
            $\hat h$ & Human-object interaction \\
            $\hat m$ & Body movement\\
            \bottomrule
        \end{tabular}}
    \label{tab:notation}
\end{table}
    
Formally, we define the notations in Tab. \ref{tab:notation}.

\subsection{Pre-Process}

For the ADT dataset, we directly obtain 3D object bounding boxes $\mathcal{O}^{3d}$, hand skeleton positions $S$, eye-gaze measurements $E$, camera poses, and egocentric video frames, and the center $c_i$ of the objects can be calculated using the 3D bounding boxes. In contrast, the Ego-Exo4D dataset does not provide explicit 3D bounding boxes, making it difficult to localize objects in 3D space. To address this issue, we leverage the annotated interaction timestamps and approximate an object's 3D position $c_i$ using the mean hand-skeleton position during the corresponding interaction interval. When both hands are involved, the average position of the two hand skeletons is adopted as the proxy for the object position. Furthermore, we extract keystep information from the atomic-description annotations in the Ego-Exo4D dataset to support our downstream analysis.

\subsection{Toolset for 3D Analysis}

\noindent\textbf{Preliminary} Before introducing the proposed toolset, we outline several core definitions and notations:
\begin{enumerate}
    \item The 3D center $c_i$ of object $i$ is computed as
    \[
        c_i=\left(\tfrac{1}{2}(o_{i,1}^{3d}+o_{i,2}^{3d}),\ \tfrac{1}{2}(o_{i,3}^{3d}+o_{i,4}^{3d}),\ \tfrac{1}{2}(o_{i,5}^{3d}+o_{i,6}^{3d})\right),
    \]
    where $o_i^{3d}\in\mathbf{R}^6$ denotes the bounding-box coordinates.
    \item The camera pose is represented by the transformation matrix $T_s^c=T_s^d \times T_d^c$, where
    \[
        T=\begin{bmatrix} R & t \\ 0 & 1 \end{bmatrix}\!,\quad 
        R\in\mathbf{R}^{3\times 3},\ t\in\mathbf{R}^{3}.
    \]
    \item The camera center $C$ corresponds to the translation component of $T_s^c$.
    \item For angular reasoning in the world coordinate system, we discretize directions into eight canonical categories: \textit{front}, \textit{back}, \textit{left}, \textit{right}, \textit{front-left}, \textit{front-right}, \textit{back-left}, and \textit{back-right}.
\end{enumerate}

\noindent\textbf{Occupancy Map Generator} The Occupancy Map Generator construct a navigation map $\mathcal{M}$ from the 3D bounding boxes $\mathcal{O}^{3d}$ observed in the last frame $x_T$ to distinguish free and occupied regions for obstacle checking. Concretely, each box is projected onto the ground plane, convex hulls are computed for the projected footprints, regions enclosed by those hulls are marked as obstacles, and the interior of the outermost hull is treated as the nominal navigable area.

\noindent\textbf{Exploration Path Generator} Given the goal object $G$ and the observation video $\mathcal X$, we can compute the center $c_i$ of $G$ and obtain the camera center $C$ from the camera pose in the last frame of $x_T$. Then the Exploration Path Generator discretizes $\mathcal{M}$ into a 2D grid, projects the start position $p_0=C$ and the goal position $p_K=c_i$ onto that grid, and runs an 8-connected A* search algorithm with direction-change penalties and diagonal-cut constraints to produce a feasible path. The resulting feasible path is represented as a sequence of waypoints, and each pair of adjacent waypoints defines a step $\hat{s}_i$. Note that we intentionally avoid using the actual human trajectory for navigation-step generation, as human motion exhibits high stochasticity and is difficult for MLLMs to reliably interpret.

\noindent\textbf{Spatial Calculator} The Spatial Calculator contains two subtools: the \textit{Distance Calculator} and the \textit{Direction Calculator}.  
The Distance Calculator projects the camera center $C$ and object centers $c_i$ into a unified world coordinate frame and computes Euclidean translation distances between queried pairs (e.g., between objects $i$ and $j$).  
The Direction Calculator computes the angle between the camera’s forward direction and the vector from $C$ to a target $G$, both projected onto the bird’s-eye-view (BEV) plane. It first extracts the camera-plane normal from $T_s^c$, projects both this normal vector and the vector from $C$ to $c_i$ into the $xOy$ plane, and then computes the resulting angle~$\theta$.

\noindent\textbf{Gaze Parser} 
The Gaze Parser converts 2D eye-tracking points $E$ into 3D gaze rays in the world coordinate system. These rays differ fundamentally from the camera-plane normal.  
For the ADT dataset, given 3D bounding boxes $o_i^{3d}$, the parser checks whether the gaze ray in future frames intersects any of the six faces of $o_i^{3d}$, while ensuring that the corresponding object appears in the last observation frame $x_T$. If multiple intersections exist, the closest one to the camera center is selected.  
For the Ego-Exo4D dataset, the parser first selects an appropriate future frame as ground truth, inserts a marker at the eye-gaze landing position, and uses an MLLM to identify the corresponding object. Using the geometric functions above, the parser returns the intentionally interacted object (and the intersection point for ADT). If a goal object is already provided, the parser instead outputs the orientation angle required to view the object.

\noindent\textbf{Affordance Detector} 
The Affordance Detector determines whether a target object will be interacted with by the observer in future frames. It operates based on three types of $\hat{h}$, described as follows:

\begin{itemize}
    \item When $\hat h$ is \textit{afford}: 
    For the ADT dataset, an object $i$ is considered to be interacted with if at least one of the following criteria is satisfied: (1) its average velocity exceeds $0.05\,\mathrm{m/s}$, or (2) the hand-skeleton position from the set of skeletons $S$ lies inside the 3D bounding box $o_i^{3d}$. The average velocity is computed as the translation distance divided by the time difference between the corresponding timestamps.  
    For the Ego-Exo4D dataset, we pre-process the timestamps of annotated interaction keysteps. The Detector then checks whether future frames contain such keysteps and selects an appropriate future frame accordingly.  
    After this determination, the Detector returns the direction and distance from the observer to the goal object in the last frame $x_T$ of the observation segment $\mathcal{X}$, using the direction and distance computation modules described earlier.

    \item When $\hat h$ is \textit{place}: 
    The Detector computes the direction from the object's current center position $c_i$ in the last observation frame $x_T$ to its predicted position in the designated future frame. It additionally ensures that the placement location is visible within the observation video $\mathcal{X}$.

    \item When $\hat h$ is \textit{action}: 
    For the Ego-Exo4D dataset, the future frame is directly provided by interaction timestamps in the annotations. The Detector uses the camera pose of the last observation frame $x_T$ and that of the future frame to compute the turn angle within the coordinate system of the camera at $x_T$. The final output follows the same format as described above.
\end{itemize}

\noindent\textbf{Keystep Extraction Tool} The Keystep Extraction Tool returns the textual keysteps in the observation video $\mathcal X$ including the interactive objects, the observer, and the interaction names from our pre-processed keystep data.

\noindent\textbf{Chain Constructor} The Chain Constructor obtains possible chains of steps and the direction between the steps. First, the Constructor obtains the processed textual keysteps from the Keystep Extraction. Then, it calculates the directions between the steps. More precisely, the \textit{direction} is the direction between the adjcent pair of waypoints in the coordinate system of camera pose in the last frame $x_T$ of the observation video. Regarding it as the basically correct chain, the tool provides several possible correct chains using multi-modal large language models.

\subsection{Toolset Usage}
In a nutshell, the 3D proximity ground truth for a given input clip sampled for each task type is constructed for each as follows:
\begin{itemize}
    \item \textbf{Intention}: The agent invokes the Spatial Calculator to estimate how the camera wearer adjusts head orientation toward the goal or directs gaze, as inferred by the Gaze Parser.
    \item \textbf{Exploration}: The agent samples a valid goal $G$ based on visibility checks and adopts the Occupancy Map Generator and Exploration Path Generator to obtain a path composed of steps $\hat{s}$ including a series of waypoints, each providing the distance and discrete direction for exploration.
    \item \textbf{Exploitation}: 
    The agent utilizes an affordance detector to identify 
    which part of the object $G$ the observer is grasping in the anticipation frame, 
    where the observer will place the object $G$, 
    and which direction the observer will move to interact with the object $G$. Which of these three types is given by $\hat h\in\{\textit{afford},\textit{place},\textit{action}\}$ specifically.
    \item \textbf{Chain of Actions}: Specifically, the agent employs the Keystep Extractor to extract key action steps and their 3D spatial locations from long video segments, and to identify the key actions toward the common goal $G$ based on future observations. It then employs an LLM to construct a set of all possible ordered combinations of key steps leading toward the same goal. Finally, The agent calls the Chain Constructor to generate a complete set of possible answers by calculating the spatial relationships among the ordered combinations of key steps.
\end{itemize}

\subsection{Post-Processing}

The proximity measurements include both approximate transformation and relative relationships. We discretize the transformation into intervals that are interpretable by humans. For spatial relationships, we convert the 3D directions into eight discrete orientations projected onto a specified plane. When constructing the candidate sets, we prompt the VLM to generate hard-negative distraction options. However, we provide specific instructions to ensure that these distractions do not rely on minor differences that are unsolvable even for humans.

We also conduct careful human verification to ensure both the validity (whether the questioned object is visible in the video clip and whether the positions we pre-process can approximate the real coordinates), answerability (whether the questions can be answered with the provided video clips) and accuracy (correctness of the answers) of the ground truth. For the \emph{Chain of Actions} task, we perform a thorough examination of all possible answer sets generated by the agent. To ensure that the question-answer pairs are contextually rich, accurate, and reflective of real-world egocentric interactions, we verified the data and removed the samples that failed our quality criteria, yielding the final benchmark.

\section{Additional Analysis on Experiments}
\label{sec:exp}
In this section, we provide additional analysis of the experiments conducted on our benchmark. Among the four tasks, Chain of Action poses a particularly significant challenge to existing MLLMs, especially when compared with human performance. In addition to the inherent difficulty of multi-step reasoning over extended temporal sequences, we observe that current models, especially open-source ones, struggle with instruction following when the input context becomes substantially longer. Recall that this task requires selecting from 10 candidate actions, which further increases the burden on the model's ability to process lengthy inputs.

Regarding the other three tasks, we observe that the Exploitation task is relatively easier for both humans and models, as it requires a much shorter temporal reasoning window. Another interesting finding is that humans are markedly better at interpreting relative spatial relationships, which naturally aligns with how people describe object locations in daily life. For existing models, estimating approximate distance appears slightly easier than identifying relative spatial relationships, since the latter requires the model to correctly infer and apply an appropriate coordinate reference.

\section{Training Details on Domain-specific Tuning}
\label{sec:train}

For all fine-tuning experiments in this work, including the
cross-category experimental setting and the cross-dataset experimental setting, we fine-tune Qwen2.5-VL-7B-Instruct with a
rank-8 LoRA adapter (\textit{target = all layers}) using the
llamafactory framework. Training is performed with bfloat16
precision, AdamW optimizer, cosine learning-rate scheduling with peak
learning rate $5\times10^{-5}$, three epochs, no warm-up, and max
gradient norm 1.0. We use an effective batch size of 16 (per-device
batch size of 2 with 8 gradient accumulation steps). FlashAttention is
enabled automatically, and both the vision tower and multimodal projector
remain frozen. 

\noindent\textbf{Cross-category fine-tuning.}
We fine-tune the model separately using 800 training examples per
category (\textit{Intention}, \textit{Exploration}, and
\textit{Exploitation}) generated from our Agentic Data Engine, allowing us to assess how specialization
on one reasoning type transfers across others.

\noindent\textbf{Cross-dataset fine-tuning.}
We additionally train the model using 1,200 QA samples from each source
dataset: ADT~\cite{pan2023aria} and EgoExo4D~\cite{Grauman2023EgoExo4DUS}.
This setting evaluates whether dataset-specific learning improves
generalization to unseen egocentric data distributions.

\section{Additional Visualization}
\label{sec:vis}

\begin{figure*}
    \centering
    \includegraphics[width=1\linewidth, keepaspectratio]{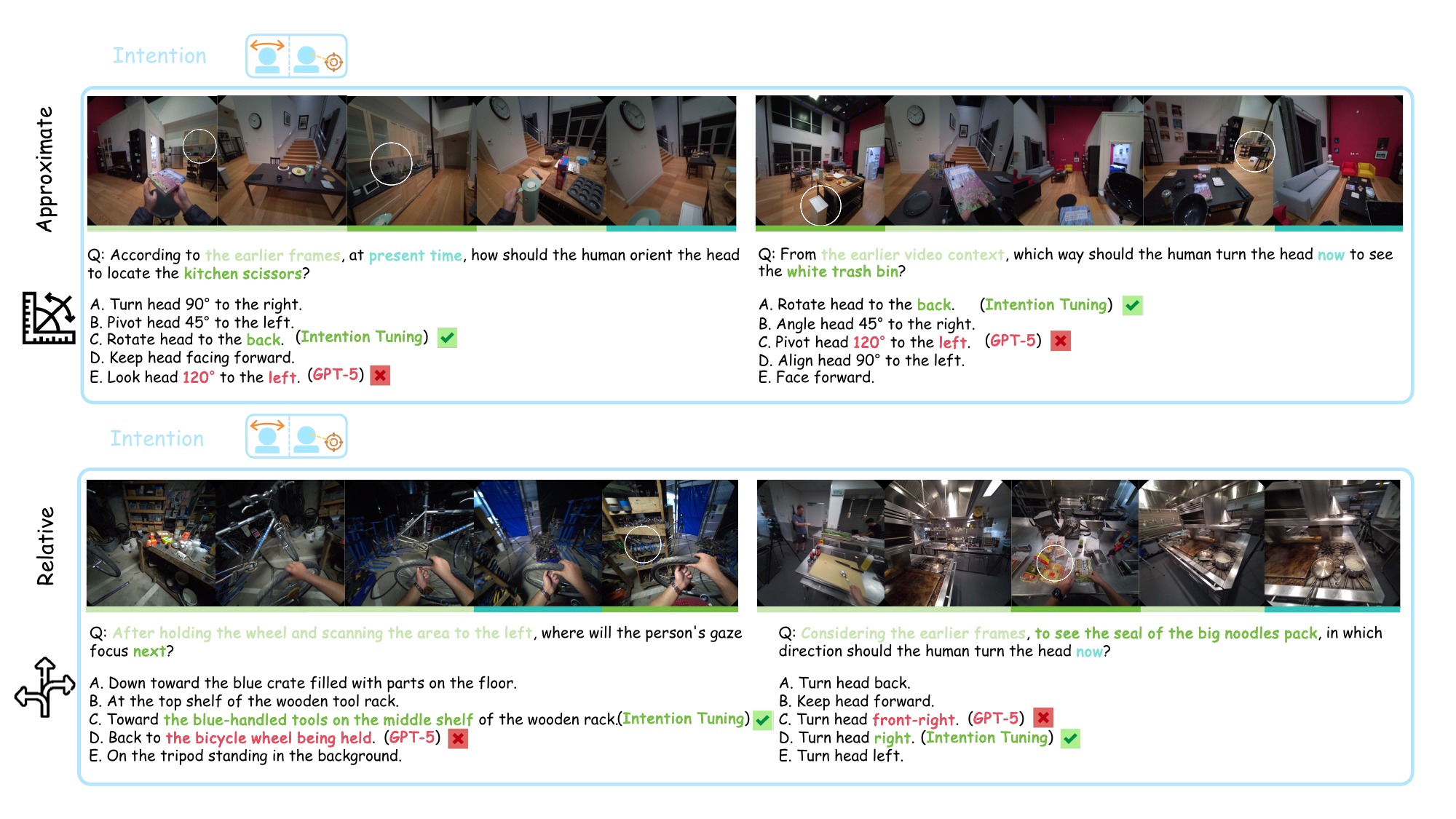}
    \vspace{-2em}
    \caption{\textbf{Visual examples of model performance on EgoProx's \textit{Intention} task.} 
We show cases where the intention-tuned model outperforms the proprietary GPT-5 model.}
    \label{fig:vis_intention}
\end{figure*}

\begin{figure*}
    \centering
    \includegraphics[width=1\linewidth, keepaspectratio]{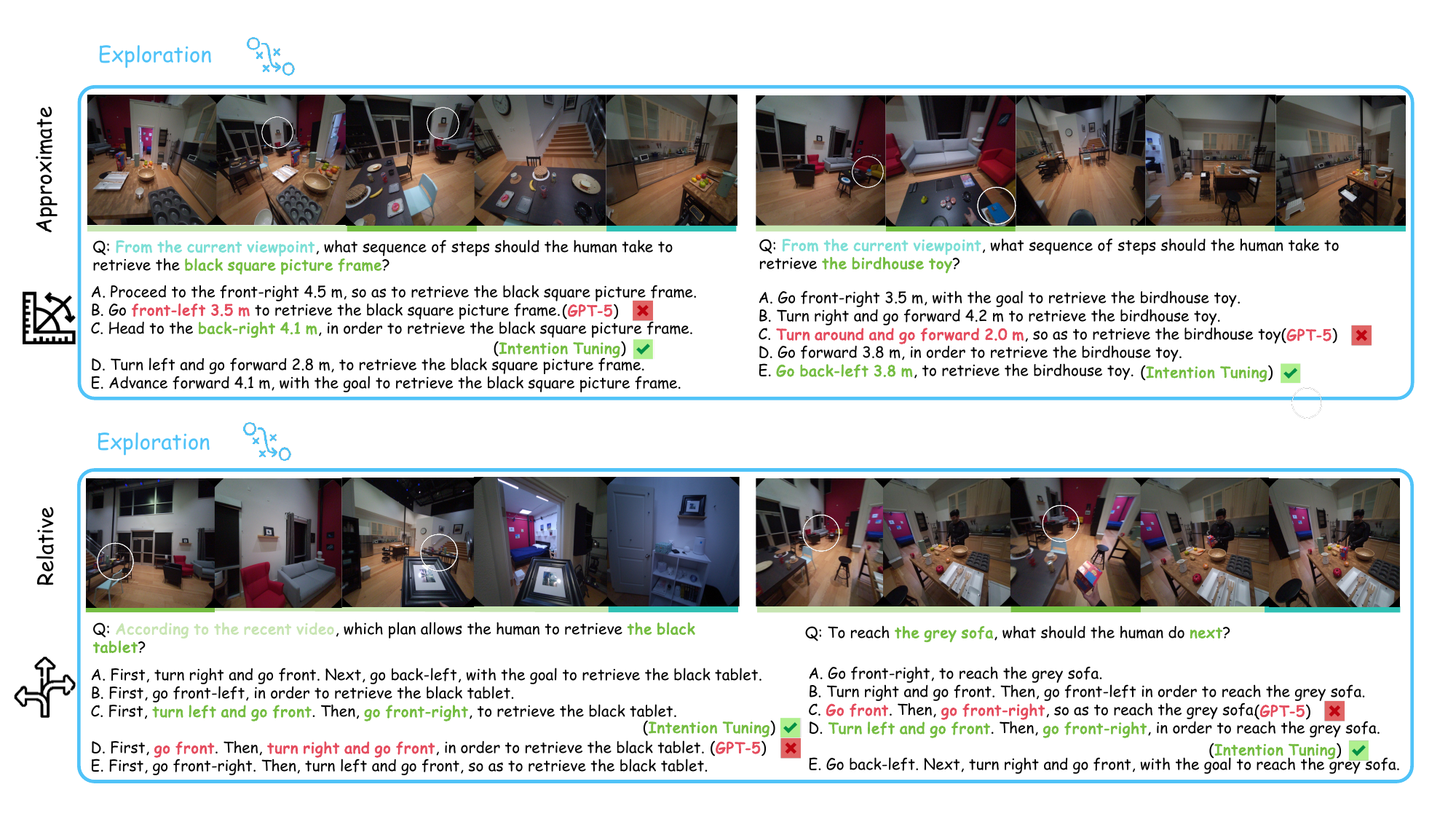}
    \vspace{-2em}
     \caption{\textbf{Visual examples of model performance on EgoProx's \textit{Exploration} task.} 
We show cases where the intention-tuned model outperforms the proprietary GPT-5 model.}
    \label{fig:vis_exploration}
\end{figure*}

\begin{figure*}
    \centering
    \includegraphics[width=1\linewidth, keepaspectratio]{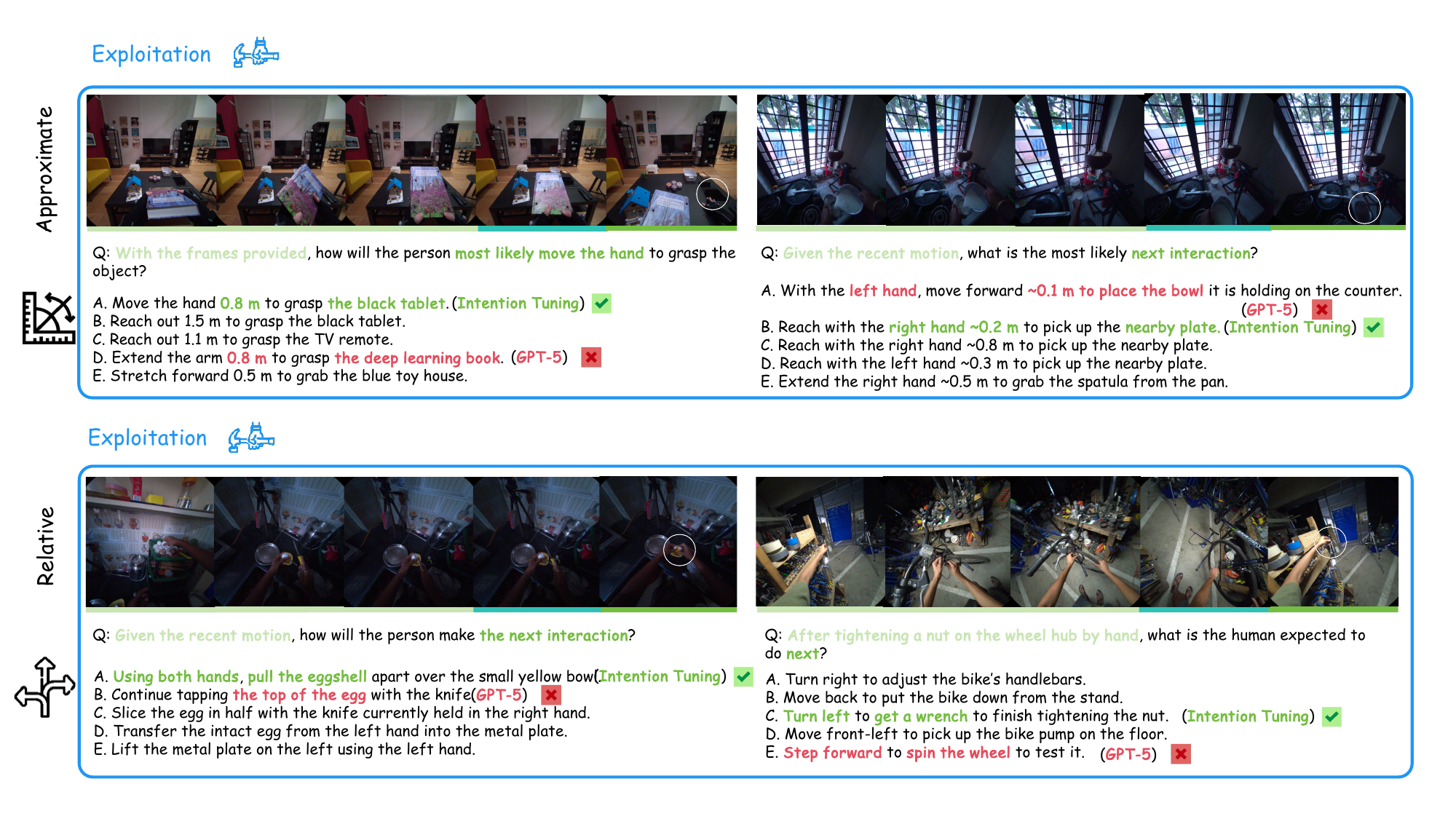}
    \vspace{-2em}
     \caption{\textbf{Visual examples of model performance on EgoProx's \textit{Exploitation} task.} 
We show cases where the intention-tuned model outperforms the proprietary GPT-5 model.}
    \label{fig:vis_exploitation}
\end{figure*}

\begin{figure*}
    \centering
    \includegraphics[width=1\linewidth, keepaspectratio]{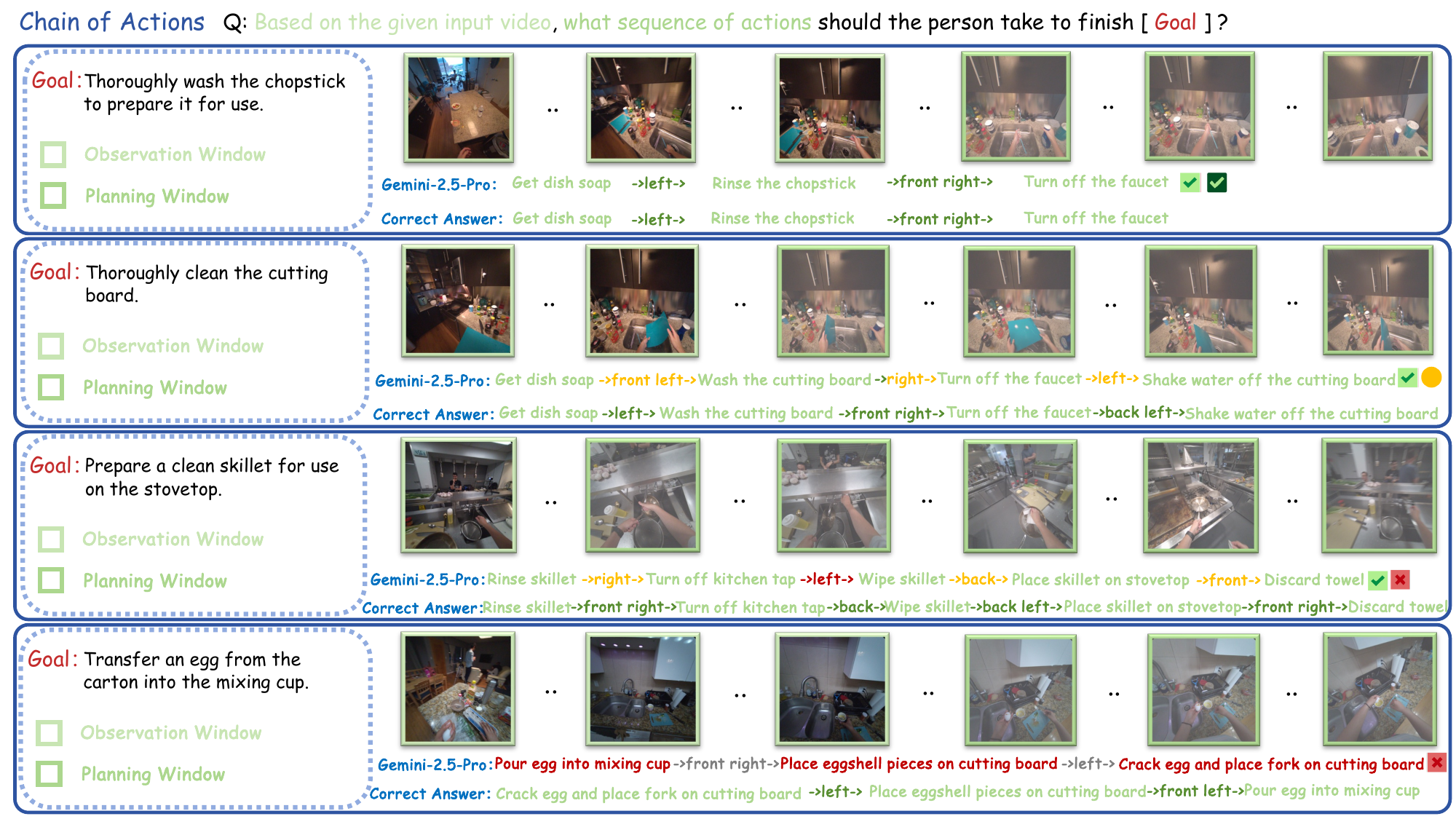}
    \vspace{-2em}
    \caption{\textbf{Visual examples of model performance on EgoProx's \textit{Chain of Actions} task.} 
We show representative cases illustrating the performance of Gemini-2.5-Pro.}
    \label{fig:vis_chain}
\end{figure*}

We provide additional visual examples to illustrate model behaviors across different reasoning tasks in \textbf{EgoProx}.
In Fig.~\ref{fig:vis_intention}, Fig.~\ref{fig:vis_exploration}, and Fig.~\ref{fig:vis_exploitation}, we showcase cases where the intention-tuned model generates more accurate and task-aligned answers compared to the proprietary GPT-5 model across the \textit{Intention}, \textit{Exploration}, and \textit{Exploitation} task categories. These examples highlight improvements in egocentric 3D Proximity reasoning after task-aware fine-tuning.

For the \textit{Chain of Actions} setting, Fig.~\ref{fig:vis_chain} illustrates representative model behaviors using Gemini-2.5-Pro. Unlike the other task types, which are multiple-choice, this task requires structured reasoning: the model must generate an ordered sequence of 3–5 action steps from a set of 10 candidates and additionally infer the spatial relationship between consecutive steps. This aligns with the formulation described in Sec.5.1, where an answer consists of a node sequence and corresponding spatial edges. To summarize model outcomes, we group examples into four types: fully correct (correct actions and spatial relationships), correct action sequence with spatial relationships correct under relaxed tolerance, correct action sequence but incorrect spatial relationships, and incorrect action sequence. These qualitative categories directly correspond to the quantitative metrics reported in main paper Table 2\&3, namely \emph{Act-Acc}, \emph{Rel-Acc-S}, and \emph{Rel-Acc-L}.

\section{Limitations}
\label{sec:limitation}

A limitation of the EgoProx benchmark lies in the coverage of egocentric scenarios. Similar to most existing egocentric datasets, our current benchmark is primarily built around indoor daily activities, which means certain environments and interaction types remain underrepresented. This reflects a common bottleneck in large-scale egocentric data collection rather than a limitation of our task design. As part of future work, we plan to further diversify EgoProx by incorporating outdoor activities and other less frequent yet representative scenarios, either through new targeted data collection or through curated web-scale egocentric videos from sources such as CommonCrawl.

One limitation of our agent-based pipeline is its reliance on video metadata, such as camera pose, 3D bounding boxes, for extracting accurate 3D information. While these annotations enable precise and scalable construction of proximity ground truth, they also limit the applicability of our pipeline to datasets that provide such metadata. As future work, we plan to integrate learned 3D perception modules, for example VGGT~\cite{wang2025vggt}, which would allow the pipeline to operate on more diverse egocentric videos without requiring pre-existing geometric annotations.

A third limitation relates to the scope of model comparisons. Following the protocol of prior Spatial AI benchmarks~\cite{Yang2024ThinkingIS,Yang2025MMSIBenchAB}, we primarily report results from prevailing general-purpose MLLMs rather than specialized spatial reasoning models. Several recent works~\cite{wu2025spatial,daxberger2025mm} have introduced architectures explicitly designed for spatial understanding, but many of these focus on generic 3D scenes or simulated environments rather than egocentric scenarios, making direct comparison less aligned with our benchmark’s goals. To maintain consistency and fairness with existing evaluation practices, we therefore do not include those models in our main results. In future work, we plan to develop more advanced spatially grounded MLLMs tailored for egocentric perception and provide comprehensive comparisons against both general-purpose and spatial-specialized models on the EgoProx benchmark.

\section{Prompt Template for Evaluation}
\label{sec:prompt}

In our experiments, incorporating a chain-of-thought style prefix leads to slightly improved performance, which is consistent with findings in existing works. We further observe that providing brief examples or explicit instructions improves the parsing success rate.

For the \textit{Intention}, \textit{Exploration}, and \textit{Exploitation} tasks in our \textbf{EgoProx} benchmark, we employ a unified prompt template for evaluation. In contrast, the \textit{Chain of Actions} task differs substantially in reasoning structure and temporal planning complexity; therefore, we adopt a separate and specialized prompt template for this task. Moreover, because the \textit{Chain of Actions} task involves varying reasoning horizons, we further provide multiple prompt variants corresponding to different action lengths.

\clearpage
\begin{tcolorbox}[
    colback=gray!2,
    colframe=gray!40,
    colbacktitle=orange!15,
    coltitle=black,
    title={Evaluation Prompt for \textit{Intention}, \textit{Exploration}, \textit{Exploitation} Tasks in  Egoprox (Without Chain-of-Thought)},
    fonttitle=\bfseries,
    arc=2pt,
    boxrule=0.3pt,
    left=6pt,right=6pt,top=6pt,bottom=6pt,
    width=\textwidth,
    breakable,
]
\scriptsize\ttfamily
\setstretch{1.2}  
\textbf{System: }\\
You are an expert in spatial reasoning, path planning, and human intention and behavior prediction. \\
You will be given a sequence of continuous first-person video frames, a question, and multiple-choice options. \\
The video is captured from the camera wearer’s own egocentric viewpoint, meaning that "the person" or "the human" mentioned in the question refers to the camera wearer. \\
All spatial directions (front, back, left, right, and their diagonals) are defined in this egocentric  viewpoint. \\
Your task is to analyze the visual content from this first-person perspective, reason about the scene in relation to the question, and select the correct answer from the provided options. 

\medskip
\textbf{User: }\\
{\ttfamily [Frame 1]}\\
{\ttfamily [Frame 2]}\\
{\ttfamily [Frame 3]}\\
...\\
{\ttfamily [Frame 8]}\\

Question: [Question text] \\

A. [Option A text] \\
B. [Option B text] \\
C. [Option C text] \\
D. [Option D text] \\
E. [Option E text] \\

Choose the most appropriate option. The selected option letter in your answer must be enclosed in angle brackets (\verb|<>|). \\

\end{tcolorbox}

\begin{tcolorbox}[
    colback=gray!2,
    colframe=gray!40,
    colbacktitle=orange!15,
    coltitle=black,
    title={Evaluation Prompt for \textit{Intention}, \textit{Exploration}, \textit{Exploitation} Tasks in Egoprox (With Chain-of-Thought)},
    fonttitle=\bfseries,
    arc=2pt,
    boxrule=0.3pt,
    left=6pt,right=6pt,top=6pt,bottom=6pt,
    width=\textwidth,
    breakable,
]
\scriptsize\ttfamily
\setstretch{1.2}  
\textbf{System: }\\
You are an expert in spatial reasoning, path planning, and human intention and behavior prediction. \\
You will be given a sequence of continuous first-person video frames, a question, and multiple-choice options. \\
The video is captured from the camera wearer’s own egocentric viewpoint, meaning that "the person" or "the human" mentioned in the question refers to the camera wearer. \\
All spatial directions (front, back, left, right, and their diagonals) are defined in this egocentric  viewpoint. \\
Your task is to analyze the visual content from this first-person perspective, reason about the scene in relation to the question, and select the correct answer from the provided options. \\
Output format: Your final line must be: The correct answer is <>.\\
Example:\\
(Reasoning...)\\
The correct answer is <B>

\medskip
\textbf{User: }\\
{\ttfamily [Frame 1]}\\
{\ttfamily [Frame 2]}\\
{\ttfamily [Frame 3]}\\
...\\
{\ttfamily [Frame 8]}\\

Question: [Question text] \\

A. [Option A text] \\
B. [Option B text] \\
C. [Option C text] \\
D. [Option D text] \\
E. [Option E text] \\

Think step by step. \\
Choose the most appropriate option. The option letter in your answer should be enclosed in angle brackets (\verb|<>|). \\
Finally, end your answer with: The correct answer is \verb|<>|.

\end{tcolorbox}

\clearpage
\begin{tcolorbox}[
    colback=gray!2,
    colframe=gray!40,
    colbacktitle=orange!15,
    coltitle=black,
    title={Evaluation Prompt for \textit{Chain of Actions} Task in \textbf{EgoProx}} (Three Actions),
    fonttitle=\bfseries,
    arc=2pt,
    boxrule=0.3pt,
    left=6pt,right=6pt,top=6pt,bottom=6pt,
    width=\textwidth,
    breakable,
]
\scriptsize\ttfamily
\setstretch{1.2}

\textbf{System: }\\
You are an expert in continuous action planning and egocentric spatial reasoning. \\
You will receive: \\
(1) a short first-person video segment consisting of 8 evenly sampled frames, where the last frame is the current observation; \\
(2) a high-level task goal that you aim to accomplish; \\
(3) a set of 10 candidate keysteps, each with an integer id; \\
(4) a discrete set of 8 egocentric directions relative to the last frame: \\
\hspace*{1em}A = right,\ B = left,\ C = front,\ D = back,\ E = front-right,\ F = front-left,\ G = back-left,\ H = back-right. \\[0.5em]
You should regard yourself as the camera wearer, i.e., the person whose first-person viewpoint is shown in the video. \\
All reasoning about space, motion, and direction must be made relative to your own egocentric viewpoint as seen in the video’s last frame. \\[0.5em]
Your task: \\
1) Choose exactly three keysteps from the candidates and order them to accomplish the goal. Return their ids as \verb|[k1, k2, k3]|. \\
2) For each transition between consecutive keysteps (from the previous interaction to the next interaction), describe the egocentric movement direction relative to your viewpoint in the last frame. Return these as two direction letters from \verb|{A, B, C, D, E, F, G, H}| for step1$\rightarrow$step2 and step2$\rightarrow$step3. \\[0.5em]
All directions are defined in your egocentric frame at the last frame: moving away from you is \verb|C| (front), moving toward you is \verb|D| (back), left/right are defined with respect to your viewpoint, and diagonals are \verb|E/F/G/H|. \\[0.5em]
After completing your reasoning, directly output only the final answer in the following format (two lists, no extra text): \\
\verb|[[k1, k2, k3], [d12, d23]]| \\[0.25em]
Example outputs: \\
\verb|[[8, 7, 3], ["F", "A"]]| \\
\verb|[[10, 7, 9], ["E", "B"]]| \\

\medskip
\textbf{User: }\\
{\ttfamily [Frame 1]}\\
{\ttfamily [Frame 2]}\\
{\ttfamily [Frame 3]}\\
...\\
{\ttfamily [Frame 8]}\\[0.5em]

Goal: [Goal text] \\[0.25em]

Candidate keysteps (id: description, total = 10): \\
1: [Keystep 1 text] \\
2: [Keystep 2 text] \\
3: [Keystep 3 text] \\
... \\
10: [Keystep 10 text] \\[0.5em]

Egocentric direction candidates (relative to the last frame): \\
A: right,\ B: left,\ C: front,\ D: back,\ E: front-right,\ F: front-left,\ G: back-left,\ H: back-right. \\[0.5em]

Please analyze the video segment and the task goal, then provide your final answer directly in the format: \verb|[[k1, k2, k3], [d12, d23]]|. \\

\end{tcolorbox}

\clearpage
\begin{tcolorbox}[
    colback=gray!2,
    colframe=gray!40,
    colbacktitle=orange!15,
    coltitle=black,
    title={Evaluation Prompt for \textit{Chain of Actions} Task in \textbf{EgoProx} (Four Actions)},
    fonttitle=\bfseries,
    arc=2pt,
    boxrule=0.3pt,
    left=6pt,right=6pt,top=6pt,bottom=6pt,
    width=\textwidth,
    breakable,
]
\scriptsize\ttfamily
\setstretch{1.2}

\textbf{System: }\\
You are an expert in continuous action planning and egocentric spatial reasoning. \\
You will receive: \\
(1) a short first-person video segment consisting of 8 evenly sampled frames, where the last frame represents the current observation; \\
(2) a high-level task goal you aim to accomplish; \\
(3) a set of 10 candidate keysteps, each with an integer id; \\
(4) a discrete set of 8 egocentric directions relative to the last frame: \\
\hspace*{1em}A = right,\ B = left,\ C = front,\ D = back,\ E = front-right,\ F = front-left,\ G = back-left,\ H = back-right. \\[0.5em]

You should regard yourself as the camera wearer — the person whose first-person viewpoint is shown in the video. \\
All reasoning about space, motion, and direction must be made relative to your own body-centered frame as seen in the last frame. \\[0.5em]

Your task: \\
1) Select exactly four keysteps from the candidates and order them to accomplish the goal. Return their ids as \verb|[k1, k2, k3, k4]|. \\
2) For each transition between consecutive keysteps, describe the egocentric movement direction relative to the last frame. Return these as \verb|[d12, d23, d34]|, where each direction is a single letter from \verb|{A–H}|. \\[0.5em]

All directions are defined relative to your egocentric viewpoint in the last frame: moving away from you corresponds to \verb|C| (front), moving toward you corresponds to \verb|D| (back), left/right are determined by your viewpoint, and diagonal movements map to \verb|E/F/G/H|. \\[0.5em]

After reasoning, output only the final result in the following format (two lists, no explanation or additional text): \\
\verb|[[k1, k2, k3, k4], [d12, d23, d34]]| \\[0.25em]

Example outputs: \\
\verb|[[8, 7, 3, 9], ["F", "A", "H"]]| \\
\verb|[[10, 7, 9, 8], ["E", "B", "A"]]| \\

\medskip
\textbf{User: }\\
{\ttfamily [Frame 1]}\\
{\ttfamily [Frame 2]}\\
{\ttfamily [Frame 3]}\\
...\\
{\ttfamily [Frame 8]}\\[0.5em]

Goal: [Goal text] \\[0.25em]

Candidate keysteps (id: description, total = 10): \\
1: [Keystep 1 text] \\
2: [Keystep 2 text] \\
3: [Keystep 3 text] \\
... \\
10: [Keystep 10 text] \\[0.5em]

Egocentric direction candidates (relative to the last frame): \\
A: right,\ B: left,\ C: front,\ D: back,\ E: front-right,\ F: front-left,\ G: back-left,\ H: back-right. \\[0.5em]

Please analyze the scene and provide your final answer directly in the format:\\
\verb|[[k1, k2, k3, k4], [d12, d23, d34]]|.\\

\end{tcolorbox}

\clearpage
\begin{tcolorbox}[
    colback=gray!2,
    colframe=gray!40,
    colbacktitle=orange!15,
    coltitle=black,
    title={Evaluation Prompt for \textit{Chain of Actions} Task in \textbf{EgoProx} (Five Actions)},
    fonttitle=\bfseries,
    arc=2pt,
    boxrule=0.3pt,
    left=6pt,right=6pt,top=6pt,bottom=6pt,
    width=\textwidth,
    breakable,
]
\scriptsize\ttfamily
\setstretch{1.2}

\textbf{System: }\\
You are an expert in continuous action planning and egocentric spatial reasoning. \\
You will receive: \\
(1) a short first-person video segment consisting of 8 evenly sampled frames, where the last frame represents the current observation; \\
(2) a high-level task goal you aim to accomplish; \\
(3) a set of 10 candidate keysteps, each with an integer id; \\
(4) a discrete set of 8 egocentric directions relative to the last frame: \\
\hspace*{1em}A = right,\ B = left,\ C = front,\ D = back,\ E = front-right,\ F = front-left,\ G = back-left,\ H = back-right. \\[0.5em]

You should regard yourself as the camera wearer — the person whose first-person viewpoint is shown in the video. \\
All reasoning about space, motion, and direction must be made relative to your own egocentric viewpoint as seen in the last frame. \\[0.5em]

Your task: \\
1) Select exactly five keysteps from the candidates and order them to accomplish the goal. Return their ids as \verb|[k1, k2, k3, k4, k5]|. \\
2) For each transition between consecutive keysteps, describe the egocentric movement direction relative to the last frame. Return these as \verb|[d12, d23, d34, d45]|, where each direction is a single letter from \verb|{A–H}|. \\[0.5em]

All directions are defined relative to your egocentric viewpoint in the last frame: moving away from you corresponds to \verb|C| (front), moving toward you corresponds to \verb|D| (back), left/right are determined by your viewpoint, and diagonal movements map to \verb|E/F/G/H|. \\[0.5em]

After reasoning, output only the final result in the following format (two lists, no explanation or additional text): \\
\verb|[[k1, k2, k3, k4, k5], [d12, d23, d34, d45]]| \\[0.25em]

Example outputs: \\
\verb|[[8, 7, 3, 4, 5], ["F", "A", "E", "B"]]| \\
\verb|[[10, 7, 9, 6, 8], ["E", "B", "D", "C"]]| \\

\medskip
\textbf{User: }\\
{\ttfamily [Frame 1]}\\
{\ttfamily [Frame 2]}\\
{\ttfamily [Frame 3]}\\
...\\
{\ttfamily [Frame 8]}\\[0.5em]

Goal: [Goal text] \\[0.25em]

Candidate keysteps (id: description, total = 10): \\
1: [Keystep 1 text] \\
2: [Keystep 2 text] \\
3: [Keystep 3 text] \\
... \\
10: [Keystep 10 text] \\[0.5em]

Egocentric direction candidates (relative to the last frame): \\
A: right,\ B: left,\ C: front,\ D: back,\ E: front-right,\ F: front-left,\ G: back-left,\ H: back-right. \\[0.5em]

Please analyze the scene and provide your final answer directly in this format:\\
\verb|[[k1, k2, k3, k4, k5], [d12, d23, d34, d45]]|. \\

\end{tcolorbox}

\clearpage
\section{Prompt Template for Training}
\label{sec:train_prompt}

\begin{tcolorbox}[
    colback=gray!2,
    colframe=gray!40,
    colbacktitle=orange!15,
    coltitle=black,
    title={LoRA Instruction-Tuning Prompt in \textbf{EgoProx}},
    fonttitle=\bfseries,
    arc=2pt,
    boxrule=0.3pt,
    left=6pt,right=6pt,top=6pt,bottom=6pt,
    width=\textwidth,
    breakable,
]
\scriptsize\ttfamily
\setstretch{1.2}

\textbf{System: }\\
You are an expert in spatial reasoning, path planning, and human intention and behavior prediction. \\
You will be given a sequence of continuous first-person video frames, a question, and multiple-choice options. \\
The video is captured from the camera wearer’s own egocentric viewpoint, meaning that "the person" or "the human" mentioned in the question refers to the camera wearer. \\
All spatial directions (front, back, left, right, and their diagonals) are defined in this egocentric viewpoint. \\
Your task is to analyze the visual content from this first-person perspective, reason about the scene in relation to the question, and select the correct answer from the provided options. \\

\medskip
\textbf{User: }\\
{\ttfamily [Frame 1]}\\
{\ttfamily [Frame 2]}\\
{\ttfamily [Frame 3]}\\
...\\
{\ttfamily [Frame 8]}\\[0.5em]

Question: [Question text] \\[0.25em]

A. [Option A text] \\
B. [Option B text] \\
C. [Option C text] \\
D. [Option D text] \\
E. [Option E text] \\[0.5em]

Choose the most appropriate option. The option letter in your answer should be enclosed in angle brackets (\verb|<>|). \\

\medskip
\textbf{Assistant: }\\
The correct answer is \verb|<[Option Letter]>|: [Chosen option text]. \\

\end{tcolorbox}


\end{document}